\documentclass[journal]{IEEEtran}

\usepackage{amsmath,amsfonts}
\usepackage{algorithmic}
\usepackage{algorithm}
\usepackage{array}
\usepackage[caption=false,font=normalsize,labelfont=sf,textfont=sf]{subfig}
\usepackage{textcomp}
\usepackage{stfloats}
\usepackage{url}
\usepackage{verbatim}
\usepackage{graphicx}
\usepackage{cite}
\usepackage{hyperref}
\usepackage[nameinlink]{cleveref}
\crefname{figure}{Fig.}{Figs}
\usepackage{pifont}   
\newcommand{\cmark}{\ding{51}}%
\newcommand{\xmark}{\ding{55}}%
\usepackage{lineno}

\usepackage{scalerel} 
\usepackage{tikz} 
\usetikzlibrary{svg.path} 
\definecolor{orcidlogocol}{HTML}{A6CE39}
\tikzset{
  orcidlogo/.pic={
    \fill[orcidlogocol] svg{M256,128c0,70.7-57.3,128-128,128C57.3,256,0,198.7,0,128C0,57.3,57.3,0,128,0C198.7,0,256,57.3,256,128z};
    \fill[white] svg{M86.3,186.2H70.9V79.1h15.4v48.4V186.2z}
                 svg{M108.9,79.1h41.6c39.6,0,57,28.3,57,53.6c0,27.5-21.5,53.6-56.8,53.6h-41.8V79.1z M124.3,172.4h24.5c34.9,0,42.9-26.5,42.9-39.7c0-21.5-13.7-39.7-43.7-39.7h-23.7V172.4z}
                 svg{M88.7,56.8c0,5.5-4.5,10.1-10.1,10.1c-5.6,0-10.1-4.6-10.1-10.1c0-5.6,4.5-10.1,10.1-10.1C84.2,46.7,88.7,51.3,88.7,56.8z};
  }
}

\newcommand\orcidicon[1]{\href{https://orcid.org/#1}{\mbox{\scalerel*{
\begin{tikzpicture}[yscale=-1,transform shape]
\pic{orcidlogo};
\end{tikzpicture}
}{|}}}}

\begin{document}

\title{Sonar-based Deep Learning in Underwater Robotics: Overview, Robustness and Challenges}

\author{Martin Aubard \orcidicon{0009-0000-3070-8067}, Ana Madureira \orcidicon{0000-0002-0264-4710}, Luís Teixeira \orcidicon{0000-0002-4050-7880}, José Pinto \orcidicon{0000-0003-1224-9391}
\thanks{This is the author’s accepted manuscript, prior to copy-editing and formatting by IEEE.  
© 2024 IEEE. Personal use of this material is permitted. Permission from IEEE must be obtained for all other uses, in any current or future media, including reprinting/republishing this material for advertising or promotional purposes, creating new collective works, for resale or redistribution to servers or lists, or reuse of any copyrighted component of this work in other works.  
}
\thanks{This project has received funding from the European Union's Horizon 2020 research and innovation programme under the Marie Sklodowska-Curie grant agreement No. 956200. For further information, please visit \url{https://remaro.eu}.} 
\thanks{M. Aubard and J. Pinto are with OceanScan Marine Systems \& Technology, 4450-718 Matosinhos, Portugal (e-mails:{maubard,zepinto}@oceanscan-mst.com), A. Madureira is with INESC INOV-Lab and ISRC (ISEP/P.PORTO), 4249-015 Porto, Portugal (e-mail:{amd@isep.ipp.pt}), L. Teixeira is with Faculty of Engineering, University of Porto (FEUP), 4200-465 Porto, Portugal (email: {luisft@fe.up.pt})}}

\markboth{This is the accepted version of the article to be published in the IEEE Journal of Oceanic Engineering.}%
{M. Aubard \MakeLowercase{\textit{et al.}}: Sonar-based Deep Learning in Underwater Robotics: Overview, Robustness and Challenges}


\maketitle

\begin{abstract}
With the growing interest in underwater exploration and monitoring, Autonomous Underwater Vehicles (AUVs) have become essential. The recent interest in onboard Deep Learning (DL) has advanced real-time environmental interaction capabilities relying on efficient and accurate vision-based DL models. However, the predominant use of sonar in underwater environments, characterized by limited training data and inherent noise, poses challenges to model robustness. This autonomy improvement raises safety concerns for deploying such models during underwater operations, potentially leading to hazardous situations. This paper aims to provide the first comprehensive overview of sonar-based DL under the scope of robustness. It studies sonar-based DL perception task models, such as classification, object detection, segmentation, and SLAM. Furthermore, the paper systematizes sonar-based state-of-the-art datasets, simulators, and robustness methods such as neural network verification, out-of-distribution, and adversarial attacks. This paper highlights the lack of robustness in sonar-based DL research and suggests future research pathways, notably establishing a baseline sonar-based dataset and bridging the simulation-to-reality gap.
\end{abstract}


\renewcommand{\arraystretch}{1.5}

\section{Introduction}
\label{sec: Introduction}
\IEEEPARstart{I}{n} recent decades, the world's oceans have become a focal point for numerous subjects of interest to scientists, industries, and military organizations, including underwater archaeology \cite{b161}, maritime exploration \cite{b162}, transportation logistics, renewable energy initiatives \cite{b163}, and military applications \cite{b164}. These applications share a common challenge: conducting safe surveys, explorations, or data collection in the often unpredictable and hazardous underwater environment. Autonomous underwater vehicles (AUVs), used to collect data and perform operations underwater, have emerged as indispensable tools in addressing this challenge.
AUVs come equipped with various sensors and instruments collecting underwater data, including measuring temperature, salinity, currents and cataloging marine species and seabed structures. However, the underwater environment presents challenges for navigation and environmental understanding due to its inherent physical uncertainties. Radio waves are rapidly absorbed by water, restricting the use of radar and global positioning system (GPS) sensors. Underwater vision is limited by reduced luminosity, turbidity, and the necessity for proximity to capture clear imagery. These challenges affect traditional vision sensors, such as red-green-blue (RGB) and grayscale cameras, and light detection and ranging (LiDAR) systems due to the significant scattering and absorption of light. On the other hand, sonar sensors, relying on sound waves that propagate efficiently and consistently underwater, provide a viable alternative for underwater sensing and navigation.

\subsection{Sonars}

Sonars, classified as active or passive, offer distinct underwater exploration and monitoring methodologies. Passive sonar captures the ambient underwater environment without emitting any signals, which is primordial for furtive surveillance. In contrast, active sonar emits sound pulses reflecting on underwater objects, marine life, or the seabed. By analyzing the return signals, we can determine object distances (time between emission and reception) and identify object types (variations in sound intensity), primordial for mapping and navigation (e.g., obstacle avoidance), providing comprehensive underwater environment information. In addition, the choice of operating frequency affects sonar performance since it directly influences the system's range and resolution as well as the level of penetration into soft bottoms such as mud. Lower frequencies provide higher propagation distances but at the cost of detail, whereas high frequencies offer superior resolution but with reduced operating range. Thus, sonar data collection requires a trade-off between range and resolution. Synthetic aperture sonar (SAS) has recently addressed this limitation by providing higher resolution without sacrificing ranges, making it a promising technology for underwater exploration.
\begin{figure}[!t]
    \centering
    \noindent \includegraphics[width=0.48\textwidth]{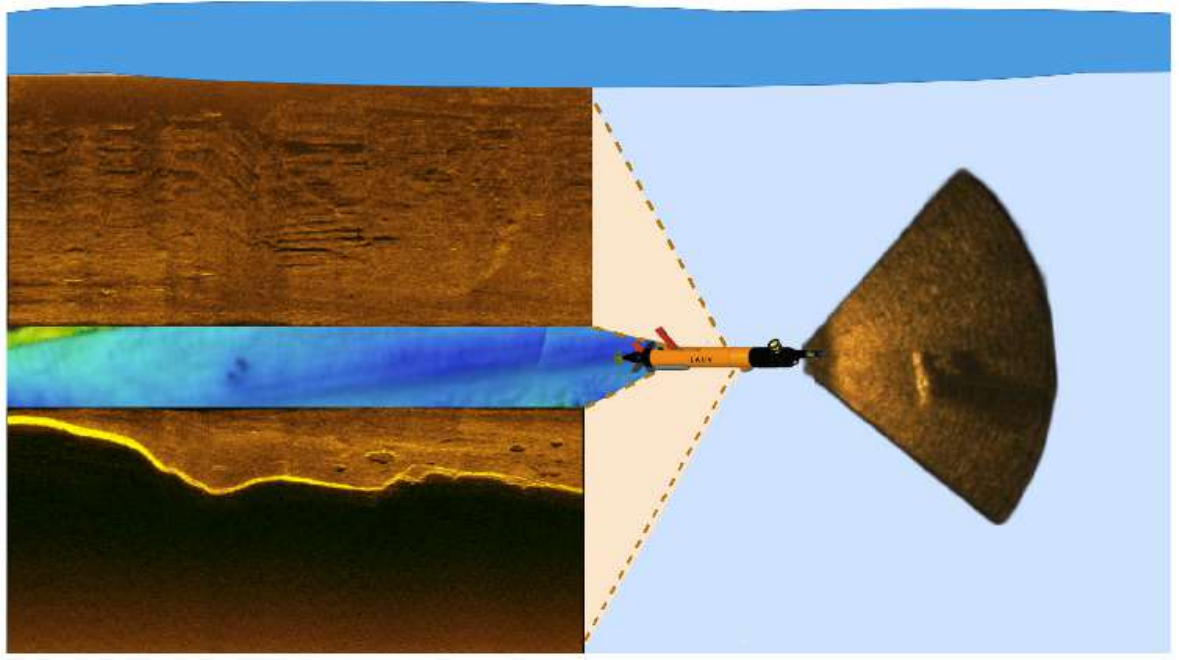}
    \caption{Sonar perception with Side Scan Sonar (SSS) \cite{aubard_2024_10528135} and Forward Looking Sonar (FLS) \cite{data_aug_dataset}. This figure represents the setup and visual information from the Multi-Beam Echo Sonar (MBES), SSS, and FLS. It shows that while the SSS provides information on past data from the port to the starboard, the FLS gives current information from the front of the AUV and the MBES from beneath the vehicle.}
    \label{Vision}
\end{figure}
Active sonars can also be categorized by configuration, such as single-beam and multi-beam types. Single-beam offers a singular sound measurement for determining altitude, aiding navigation tasks (e.g., bottom tracking). In contrast, multi-beam sonar, including side-scan sonar (SSS) and multi-beam echo sounders (MBES), are designed to simultaneously capture a broader range of data points. MBES provides a bathymetric mapping from beneath the vehicle, whereas SSS focuses on generating high-resolution seafloor imagery on both sides of the vehicle (port and starboard). In addition, forward-looking sonar (FLS), whether single-beam or multi-beam, maps the area directly ahead of the vehicle. A single-beam FLS might only provide distance measurements to upcoming obstacles, whereas a multi-beam FLS offers a detailed representation of the area in front of the vehicle, enhancing obstacle avoidance and navigation. \cref{Vision} represents three sonar sensors: an SSS on both port and starboard sides, an MBES beneath, and a multi-beam FLS in front of the vehicle. Due to the position of the SSS, the usual SSS image lacks information between the two transducers, which is called the nadir gap; however, in this graphical representation, the MBES fills the nadir gap.

\subsection{Deep Learning and Robustness}
Artificial Intelligence (AI) has emerged as a crucial solution to enhance the performance and capabilities of AUVs. AI's ability to process and interpret large datasets, identify patterns, and make informed predictions can significantly improve AUV navigation, collision avoidance, and real-time decision-making \cite{b165}. Over the past decade, Machine Learning (ML) and Deep Learning (DL) techniques have been applied to improve feature detection in data collected by AUVs, such as identifying mines \cite{b166} or shipwrecks \cite{b167}.
Traditionally, ML and DL methods have been used in post-processing vision-based data, where collected data is processed offline to detect features. This approach reduces the time needed for target identification, which typically requires domain expertise. However, due to the time-consuming nature of offline processing, AUVs cannot interact with potential underwater objects while surveying. Recently, the focus has shifted towards integrating DL algorithms directly onboard AUVs to enable real-time interaction with the underwater environment, enhancing responsiveness to environmental changes and improving the autonomy of AUVs \cite{aubardLSTSDL}.
This real-time approach enables AUVs to adapt their navigation dynamically in response to detected objects or obstacles, streamlining survey times and improving data quality \cite{aubard2024mission}. However, relying on DL models for interpreting environmental data in real time raises essential questions about the reliability of sonar-based DL outputs.

The central question, "\textbf{How can we rely on real-time sonar-based deep learning models?}" raises several related issues regarding the robustness of DL models and the true meaning of robustness.
Robustness refers to the ability of AI systems to handle errors or inconsistencies during their operation \cite{b188}. It is a subcategory of the broader concept of Safe AI, including Explainable AI (XAI), interpretability, privacy, and security. Safe AI aims to ensure the reliability of algorithms, vehicle safety, and the safety of surrounding environments \cite{b59}. Due to their opacity, DL algorithms are often considered 'black boxes' \cite{b184}, as their reasoning process is unclear, requiring measures to ensure expected performance while mitigating unintended consequences and potential harm \cite{b185}.
The challenge lies in balancing the benefits of AI with managing the risks of misuse or malfunction. Explainable AI (XAI) \cite{b191} provides insights into the training and implementation processes, helping users understand AI behaviors, identify issues, and improve system architecture. The EU's General Data Protection Regulation (GDPR) \cite{b190} enforces a "right to explanation," allowing individuals affected by AI decisions to request an explanation of how those decisions were made. Interpretability \cite{b186} further supports understanding the model structure, aiding in reliability assessments, diagnostics, and corrections.
While Safe AI has seen significant progress in terrestrial applications, such as autonomous driving, there has been less focus on ensuring the safety of underwater perception systems, particularly sonar-based systems. Underwater environments pose unique challenges, including unreliable GPS positioning, poor visibility for optical sensors, noisy sonar data, strong currents, and constantly changing conditions. 
Hence, implementing DL onboard AUVs raises questions about vehicle safety, where for the AUV itself, safety means ensuring that it does not collide due to unreliable model behavior, does not get lost, and collects data accurately.

\subsection{Motivations}

DL models' robustness examination aims to consistently ascertain their ability to make accurate predictions under all conditions. Furthermore, it is essential to understand the factors that could lead to erroneous sonar-based DL outputs. The data-driven nature of DL models requires relying on the integrity and relevance of training data, which brings up the following critical considerations: 1) Adequate data volume in the training set is essential for comprehensive learning and recognition abilities upon deployment. 2) The relevance of data to the operational environment is vital, prompting questions about what constitutes the resemblance of data to the deployment context regarding environmental and sensor data distributions. Those considerations are even more critical for sonar-based datasets due to their open-source accessibility limitation.  
Thus, if the available data prove insufficient, operators face the following alternatives: 1) collecting more data (time and cost-consuming); 2) utilizing simulators for data generation, requiring high fidelity to real-world conditions; or 3) leveraging generative models to increase the dataset volume. Nevertheless, even with precise data, model-induced errors remain possible, underscoring the significance of neural network verification as a discipline seeking to mathematically validate model reliability within defined noise thresholds. However, the current state of neural network verification tools reveals limitations in applicability across tasks and constraints related to model size or activation functions. Adversarial attack and out-of-distribution detection emerge, focusing on identifying and mitigating noise that could deceive the DL model, emphasizing the importance of incorporating adversarial attack defenses as part of the pre-deployment model evaluation process. Sonar data is affected by various underwater noise sources, including self-noise (system-generated noise), multi-path reflections (reflected signals from the ocean surface and bottom), interference from other sonars (external interference), ambient noise from marine life and oceanic activity (environmental noise), as well as speckle noise inherent in sonar imaging (imaging artifact) \cite{Sonar-book}. Furthermore, the disparate distribution between sonar brands can also mislead DL output. Thus, those pre-deployment steps remain essential for reducing unpredicted noises potentially encountered during a mission. 
\\

This document aims to guide practitioners in deploying DL models for sonar on AUVs, emphasizing model robustness considering the uncertainties inherent in sonar data. Such model presents unique challenges that are less documented than traditional optical data, requiring specific focused handling uncertainties and ensuring reliable model performance. The following sections explore the complexities of utilizing sonar-based DL in underwater robotics and provide a comprehensive overview of improving the robustness of these applications. The objective is to demonstrate how leveraging sonar-based DL can significantly enhance AUV capabilities while ensuring the safety and success of underwater missions.
\Cref{sec: Sonar-based - Previous Surveys} reviews existing literature, positioning this work as a pioneering effort in addressing the robustness of sonar-based DL models. \Cref{sec: AI-Based Underwater Robotics Tasks} discusses essential sonar-based tasks enabled by DL, including classification, object detection, segmentation, and SLAM. Finally, \autoref{Safe AI - Sonar-Based Computational Vision DL} explores the current state of robustness in sonar-based applications, highlighting state-of-the-art open-source datasets, simulators for underwater sonar environments, and synthetic data generation techniques. It also emphasizes the emerging field of neural network verification in sonar contexts and discusses methodologies for out-of-distribution (OOD) detection, adversarial attacks, and uncertainty quantification.

\section{Sonar-based Deep Learning - Related Surveys}
\label{sec: Sonar-based - Previous Surveys}
The exploration of DL in sonar imagery for underwater applications is captured through the following insightful review papers, each delving into different aspects and methodologies in the field of sonar-based DL. Firstly, D. Neupane et al. \cite{electronics9111972} broaden the discussion by emphasizing the importance of sonars in underwater object detection and the challenges posed by the lack of accessible datasets. The article provides a structured analysis that spans sonar principles, the utility of DL over traditional AI and ML-based methods, and a detailed examination of datasets, pre-processing technologies, and DL architectures for Automatic Target Recognition (ATR), highlighting the intrinsic challenges of sonar data, such as non-homogeneous resolution and acoustic shadowing. The review underscores the need for high-quality, shared datasets to advance the field. It recommends detailed documentation of datasets, simulators for data generation, and a more nuanced approach to data augmentation for enhancing ATR in sonar-based studies.
Y. Steiniger et al.  \cite{STEINIGER2022105157} distinguishes itself by concentrating on Synthetic Aperture Sonar (SAS) and SSS data, areas not extensively covered in prior reviews. Acknowledging the foundational work by \cite{electronics9111972}, this paper narrows its focus to SSS and SAS, excluding FLS and MBES from its analysis. It rigorously compares simple convolutional neural networks (CNN) algorithms across various tasks such as feature extraction, classification, object detection, and segmentation, highlighting a comprehensive examination of over 60 publications related to SSS object detection. The paper critically notes the absence of open-source SSS image datasets, which hampers the comparability of research outcomes, and suggests data augmentation and the generation of simulated data using generative adversarial network (GAN) models as potential remedies. However, it also highlights the scarcity of shared datasets and models that could facilitate broader research collaboration.
A. Khan et al. \cite{review_2024} present a selection of cutting-edge algorithms developed over the past seven years, filling gaps that have yet to be addressed by existing surveys and catalogs the applications for underwater object detection. It provides a succinct overview of architectures, including a comparative analysis of various YOLO versions and other CNN-based models, termed "ConVNNs". 
\begin{table*}[!htbp]
\caption{Comparative Analysis between the Surveys on Sonar-based Deep Learning. This table compares the surveys on the Sonar-based Deep Learning by analyzing the date of publication (\textbf{Year}), the main topic of the survey (\textbf{Topics}), if they enumerate and compare SAO datasets (\textbf{Datasets}), if they enumerate and compare underwater simulator for sonar-based image generation (\textbf{Simu.}), if they enumerate and compare SAO DL models (\textbf{Models}), the type of sensors used (\textbf{Sensors}), the type of tasks realized by the DL models (\textbf{Tasks}), where \textit{C} stands for Classification, \textit{D} for Detection, \textit{S} for Segmentation, and \textit{Sl} for SLAM, and finally the robustness method tackle in the surveys (\textbf{Robust.}), where \textit{OOD} stands for Out-Of-Distribution, \textit{AA} for Adversarial Attack, and \textit{UQ} for Uncertainty Quantization. The \textbf{*} symbol translates a brief mention of the topics in the paper.} 
\centering
\footnotesize
\label{table:comparisonsSurveyPaper}
\begin{tabular}{l@{\hspace{.3mm}} c@{\hspace{.8mm}} c@{\hspace{.8mm}} c@{\hspace{.4mm}} c@{\hspace{.3mm}} c@{\hspace{.3mm}} c@{\hspace{.3mm}} c@{\hspace{.3mm}} c@{\hspace{.3mm}}}
\hline
\textbf{Survey} & \textbf{Year} & \textbf{Topics} & \textbf{Datasets } & \textbf{ Simu. } & \textbf{ Models} & \textbf{Sensors} & \textbf{Tasks} & \textbf{Robust.} \\    
\hline
D. Neupane et al . \cite{electronics9111972} & 2020 & ATR & \cmark & \cmark &\cmark & SSS, FLS & C, D, S & \xmark\\
Y. Steiniger et al. \cite{STEINIGER2022105157} & 2022 & ATR & \xmark & \xmark &\cmark & SSS, SAS & C, D, S & UQ \textbf{*}\\
A. Khan et al. \cite{review_2024} & 2024 & ATR & \xmark & \xmark &\cmark  & SSS, FLS & C, D & \xmark\\
B. Teng et al. \cite{doi:10.1177/1729881420976307} & 2020 & ATR & \xmark & \xmark &\cmark & SSS, FLS, RGB & C, D & \xmark \\
Y. Tian et al. \cite{doi:10.1177/1729881420936091} & 2020 & Sonar Segmentation & \xmark & \xmark & \cmark & SSS, FLS & S & \xmark \\
A. Yassir et al. \cite{YASSIR2023106790} & 2023 & Fish Identification & \xmark & \xmark & \cmark & MBES & C, S & \xmark\\
Y. Chai et al. \cite{10299606} & 2023 & Fish Identification & \cmark & \cmark \textbf{*} & \cmark & SSS, SAS, FLS & C, D, S & \xmark\\
L. Domingo et al. \cite{s22062181} & 2022 & Shoreline Surveillance & \cmark & \xmark & \cmark & Passive Sonar & C & \xmark\\
\hline
\textbf{Ours}& \textbf{2024} & \textbf{ATR} & \cmark & \cmark & \cmark & \textbf{SSS, FLS} & \textbf{C,D,S,Sl} & \textbf{ OOD,AA,UQ}\\
\hline
\end{tabular}
\end{table*}
The paper calls for a more diversified and balanced dataset, exploring deep transformer models and developing hybrid detection techniques, among other future directions.
B. Teng et al. \cite{doi:10.1177/1729881420976307} conducted a comprehensive survey of DL-based detection methods for mines and manmade targets using underwater RGB and sonar imagery. They thoroughly examine the various types of noise encountered in sonar images and discuss methods to mitigate these noises during data processing. Additionally, they evaluate several SOA models using identical datasets to benchmark performance for RGB image analysis.
Y. Tian et al. \cite{doi:10.1177/1729881420936091} work is the pioneering and sole survey paper on sonar segmentation. It meticulously reviews existing sonar segmentation literature, outlines the current challenges in the field, and proposes 12 research directions to advance sonar segmentation studies.
Expanding the scope of our analysis, we study sonar applications in fish identification and shoreline surveillance. A. Yassir et al. \cite{YASSIR2023106790} focus on fish classification and segmentation, comparing SOA DL models. In contrast, Y. Chai et al. \cite{10299606} provide an extensive survey covering fish classification, detection, segmentation, and denoising of sonar images. However, the comparison by A. Yassir et al. \cite{YASSIR2023106790} across models trained on diverse datasets complicates definitive conclusions due to the potential variability in dataset characteristics.
L. Domingo et al. \cite{s22062181} explore DL methods for shoreline surveillance by classifying underwater vessels using passive sonar. This review highlights the adaptability of DL methods to several aspects of underwater exploration and monitoring, showcasing the breadth of potential applications for sonar technology.
In synthesizing these reviews, it becomes evident that while significant strides have been made in applying DL to sonar imagery for underwater detection, the field still faces substantial limitations, such as the critical need for open-source datasets, a more granular understanding, and explanation of DL models tailored to sonar data. However, because of the need for a sonar baseline dataset, comparing all the DL models does not give concrete insight into which one is the most suited for sonar images. \autoref{table:comparisonsSurveyPaper} highlights a detailed comparative analysis of surveys on sonar-based DL models. This comparison underscores the singularity of our contribution relative to previous works in the scope of sonar-based DL. Our survey is the first to provide an in-depth comparison of current sonar open-source datasets and simulators and to address the robustnesses of sonar-based DL models, delving into critical areas such as Out-of-Distribution (OOD) detection, adversarial attacks, and uncertainty quantification. This comprehensive approach aims to provide valuable insights into the current SOA of sonar-based DL in terms of models, datasets, simulators, and methods to improve the robustness of the DL prediction.

\section{Sonar-based Deep Learning Perception}
\label{sec: AI-Based Underwater Robotics Tasks}
This section describes the principal tasks achievable by applying sonar-based DL onboard AUVs, explicitly focusing on classification, object detection, segmentation, and Simultaneous Localization and Mapping (SLAM). This discussion includes a historical overview of the model developments and highlights the SOA models for these tasks. In this paper, we consciously abstain from directly comparing the performance of published sonar-based DL models. This decision is twofold; firstly, as indicated in \autoref{sec: Sonar-based - Previous Surveys}, drawing direct comparisons across models is challenging due to the utilization of disparate datasets collected under varying conditions with different sonar equipment, most of which are not publicly accessible. Secondly, prior survey papers, including \cite{electronics9111972}, \cite{STEINIGER2022105157}, and \cite{review_2024}, have already undertaken comprehensive comparisons of sonar-based DL models published respectively in 2020, 2022, and 2024. We direct interested readers to these surveys for in-depth comparisons. Conversely, our section's objective is to furnish readers with an encompassing perspective on the assortment of models deployable for sonar-based DL tasks such as classification, object detection, segmentation, and SLAM, offering insights into the evolving landscape of DL.

\subsection{Classification and Object Detection}
Classification and object detection integrated into underwater vehicles effectively enhance the situational awareness and navigation in a complex underwater environment. Classification algorithms allow for categorizing underwater objects or features into predefined classes, which is essential for tasks like marine life monitoring, habitat mapping, and underwater archaeology. Object detection, on the other hand, extends this capability by identifying these categories and locating and tracking objects within the sonar imagery, which is crucial for obstacle avoidance, target tracking, and detailed environmental assessment.
The history of DL classification has seen significant evolution, especially with the advent of Convolutional Neural Networks (CNNs). Since the breakthrough achievement of AlexNet in 2012 \cite{alexnet}, CNNs have become a staple in computer vision tasks, including object detection, segmentation, and classification. They leverage convolutional layers to reduce image size and enhance pattern recognition capabilities effectively. AlexNet, a successor to LeNet \cite{lenet}, introduced more filter layers and demonstrated remarkable classification abilities across over a thousand classes using RGB images. ResNet \cite{resnet} further revolutionized DL by introducing residual networks with "skip connections" to combat the vanishing gradient problem, improving accuracy even as network depth increased.
In the domain of object detection, the last decade has witnessed remarkable progress. It encompasses crucial tasks like object localization and classification within images, facilitated by both one- and two-stage detectors. Two-stage detectors excel in accuracy by separately addressing localization and classification, but often at the cost of efficiency. The inception of R-CNN in 2014 \cite{rcnn}, followed by advancements like Fast \cite{b94} and Faster R-CNN \cite{b95}, highlighted the efforts of optimizing the balance between accuracy and efficiency, mainly through innovations like Region of Interest (ROIs) pooling and separate networks for region proposal predictions.
The emergence of one-stage detectors, notably through the YOLO (You Look Only Once) \cite{b96} series, underscored a pivotal shift toward real-time object detection by merging localization and classification tasks. Despite initial trade-offs in accuracy, subsequent iterations of YOLO and SSD (Single Shot Detector) \cite{b102} have improved the efficiency and effectiveness of object detection models. These models, particularly YOLOv4 and its successors, have set new benchmarks in real-time detection capabilities, addressing challenges like small object detection through innovations like FPP necks and CSPDarkNet53 backbones.
Initially introduced for Natural Language Processing (NLP), transformers \cite{b104} have broadened their application to include computer vision tasks. Their architecture, centered around the self-attention mechanism, facilitates a deeper contextual understanding of input sequences and has proven particularly effective in overcoming the limitations of sequential models like Recurrent Neural Networks (RNNs) \cite{RNN} and Long Short-Term Memory (LSTMs) \cite{b88}, offering enhanced data handling and processing capabilities. Adapting transformer technology to vision tasks, as demonstrated by DETR (DEtection TRansformer) \cite{b106} and ViT (Visual Transformer) \cite{b107}, signifies a paradigm shift in object detection, achieving unparalleled accuracy and context awareness, although with higher data and computational requirements.
\begin{figure}[!t]
    \centering
    {\begin{tabular}
    {cc}
    Forward-Looking Sonar \cite{NEVES2020112870} & Side Scan Sonar \cite{b8}\\
    \includegraphics[width=0.23\textwidth]{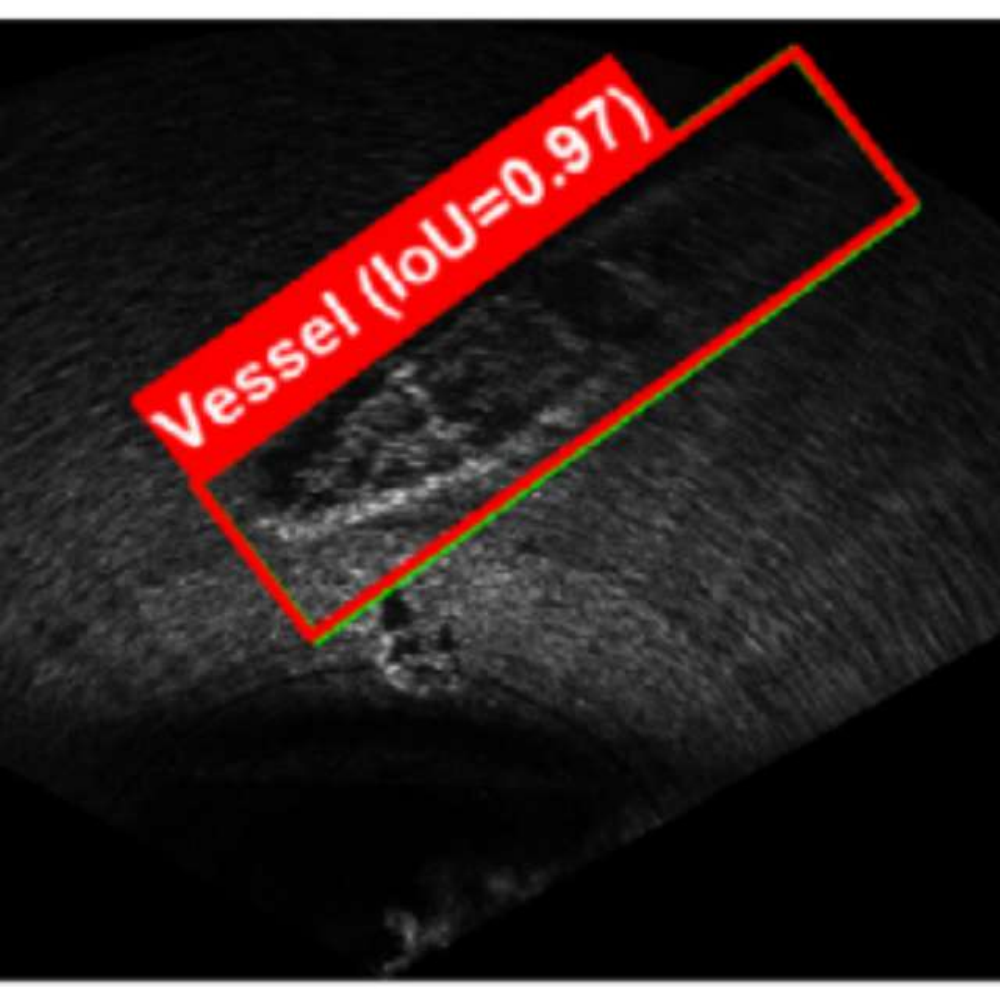} & \includegraphics[width=0.23\textwidth]{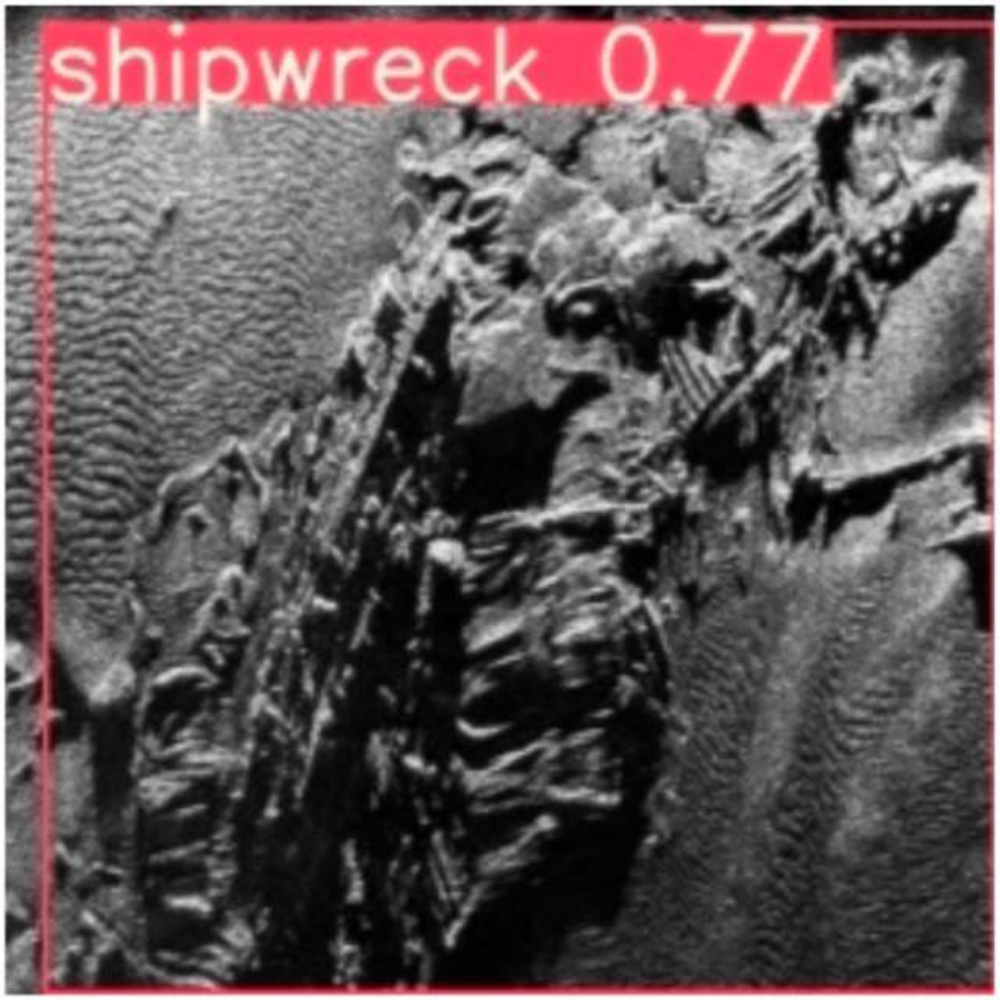}\\
    \end{tabular}}
    \caption{Samples object detection on sonar images. Those two samples show an object detection model prediction for a vessel on an FLS image and a shipwreck on an SSS image.}
    \label{sampleDetectionSonar}
\end{figure}
Due to these requirements, their applicability in underwater scenarios is limited by significant data and computational demands. These models typically require extensive datasets to achieve performance comparable to conventional CNN-based SOA, a challenging underwater detection precondition due to the scarcity of large-scale datasets. 
Nevertheless, recent efforts have explored integrating transformer layers with CNNs to leverage the transformers' superior feature extraction capabilities while mitigating their data requirements, leading to the development of models such as YOLOv5-TR \cite{b8}, where the transformer layer is located between the neck and the backbone layers, demonstrating improved performance with the complexities of underwater environments in SSS images. M. Aubard et al. \cite{Aubard-wall-Detection} compared the YOLOv5 with its transformer version YOLOv5-TR and a novel anchor-free object detection of the YOLOX \cite{b101} with an SSS wall dataset. They conclude that YOLOv5-TR improved the classic version, whereas YOLOX gives the best result. Recently, the authors proposed the YOLOX-ViT model \cite{yolox-vit} in SSS images, improving its previous version implemented by a ViT layer, and proposed a lightweight version of their model called KD-YOLOX-ViT using knowledge distillation. 
Two object detection sonar samples are represented in \cref{sampleDetectionSonar}. The left sample represents the RBoxNet rotation bounding end-to-end detector \cite{NEVES2020112870} detecting a shipwreck using an FLS. In contrast, the right sample shows the YOLOv5-TR \cite{b8} detecting shipwrecks on SSS images. Although both detections detect shipwrecks, those samples show the detection representation of shipwrecks on SSS and FLS images.  
\\

\subsection{Segmentation}
The goal of segmentation in underwater imagery is to categorize each pixel of an image into meaningful classes, which can significantly assist in tasks like habitat mapping, species identification, and monitoring underwater infrastructure or ecological changes. 
The journey of DL-based segmentation began with Fully Convolutional Networks (FCNs) \cite{FCN_segmentation}, which marked a departure from traditional patch-based classification methods by processing an entire image in a single forward pass and outputting a pixel-wise annotation map. U-Net's architecture \cite{unet}, characterized by its symmetric expanding and contracting paths, was designed to capture context and localize features effectively. This model became a blueprint for many follow-up studies, including those focusing on underwater imagery, due to its efficiency in handling small datasets with high performance, a common scenario in underwater research. As the field progressed, models like DeepLab \cite{deeplab} and PSPNet \cite{pspnet} introduced approaches to capture broader context and achieve more precise segmentation boundaries. These models enhanced the segmentation of complex scenes, where the variability in scale and appearance of objects poses significant challenges.
\begin{figure}[!t]
    \centering
    {\begin{tabular}
    {cc}
    Forward-Looking Sonar \cite{CHEN2021102691} & Side Scan Sonar \cite{YU2021102608}\\
    \includegraphics[width=0.23\textwidth]{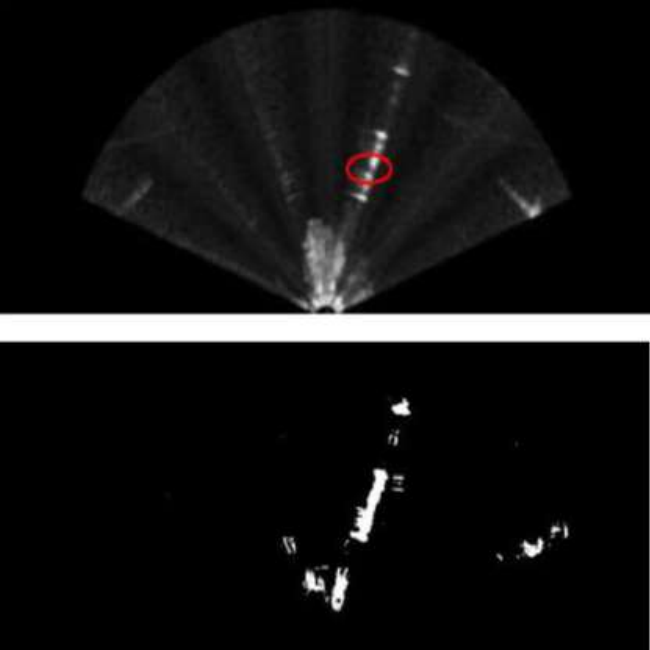} & \includegraphics[width=0.23\textwidth]{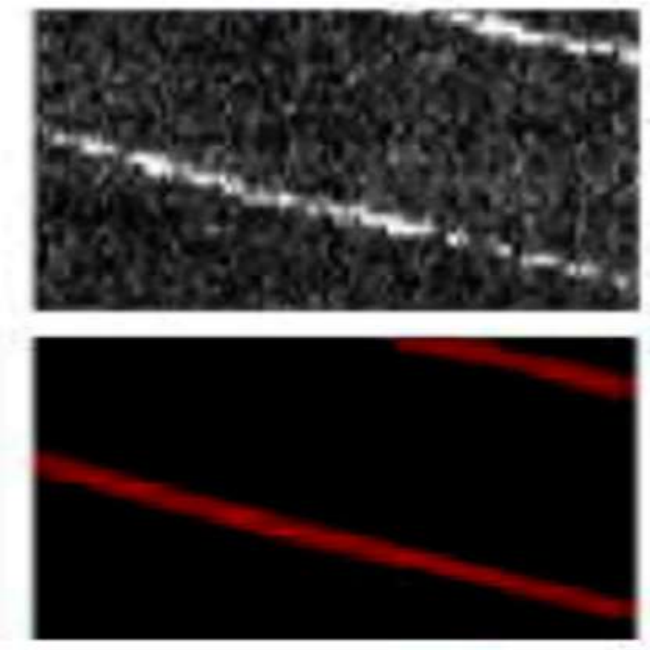}\\
    \end{tabular}}
    \caption{Samples segmentation on sonar images. Those two samples show a segmentation model prediction on FLS and SSS images.}
    \label{sampleSegmentationSonar}
\end{figure}
As for the object detection field, introducing attention mechanisms and transformers into segmentation models marked another improvement. Attention U-Net \cite{attention-unet}, for instance, adapted the U-Net architecture by incorporating attention gates, which help the model focus on relevant features while suppressing less important ones. This capability is particularly beneficial in underwater segmentation, where foreground-background contrast can be low, and objects of interest may be obscured. Models like SETR \cite{setr} and ViT-Seg \cite{vit-seg}, which leverage the transformer's ability to handle long-range dependencies, offer promising results for feature extraction of underwater segmentation into sonar images such as MiTU-Net \cite{9807763} a mix of a Transformer and U-Net used on FLS images. However, transformer-based models for feature extraction in sonars can be unstable due to unpredictable noise; J. He et al. \cite{10443635} have proposed a hybrid CNN-Transformer-HOG (Histogram of Oriented Gradient) framework for FLS segmentation to address this challenge, outperforming the previous CNN and CNN-Transformer-based sonar segmentation. Most underwater segmentation studies predominantly utilize RGB images, reflecting a broader trend within computer vision \cite{alvareztunon2024subpipe}, \cite{b17}. 
However, there is a notable gap in the availability of datasets derived from sonar imaging. This scarcity is compounded by the significant time and resources required for dataset annotation, particularly for segmentation tasks.
Segmentation, by its nature, demands detailed pixel-wise labeling, which is considerably more time-consuming and labor-intensive than the bounding box annotations used in object detection. Each image must be meticulously analyzed to ensure accurate classification of all pixels, which can be incredibly challenging in underwater environments where distinguishing between different elements can be complicated due to poor visibility, sound reflection/refraction, overlapping objects, and variable lighting conditions. The substantial annotation effort required for segmentation datasets sometimes aligns with the limited resources and time available to research teams. 
\cref{sampleSegmentationSonar} represents two segmentation samples; the one on the left represents a saliency segmentation method for pipeline recognition on FLS \cite{CHEN2021102691}, whereas the right sample is an SSS image where a combined residual and recurrent CNN called R²CNN is applied on a fishing net dataset \cite{YU2021102608}.

\subsection{Simultaneous Localization and Mapping (SLAM)}
The final subfield in computer vision addressed in this section is Simultaneous Localization and Mapping, commonly known as SLAM \cite{b156}. In contrast to detecting and localizing objects, SLAM focuses on mapping an unknown environment while simultaneously tracking the robot's position. SLAM algorithms are crucial in robotics and autonomous vehicle navigation, especially in scenarios where GPS is unreliable or unavailable, such as underwater. It usually employs probabilistic algorithms such as Kalman filters \cite{b173} or Graph SLAM \cite{b172} to represent uncertainty in both the map and the robot's position estimate. These algorithms can handle several types of sensor data, including vision (optical and sonar), range, or odometry measurements, and fuse these measurements to create a consistent environment map while estimating the robot's pose.
\begin{figure}[!t]
    \centering
    {\begin{tabular}
    {cc}
    MSIS \cite{9705938} & Side Scan Sonar \cite{Zhang_2023}\\
    \includegraphics[width=0.23\textwidth]{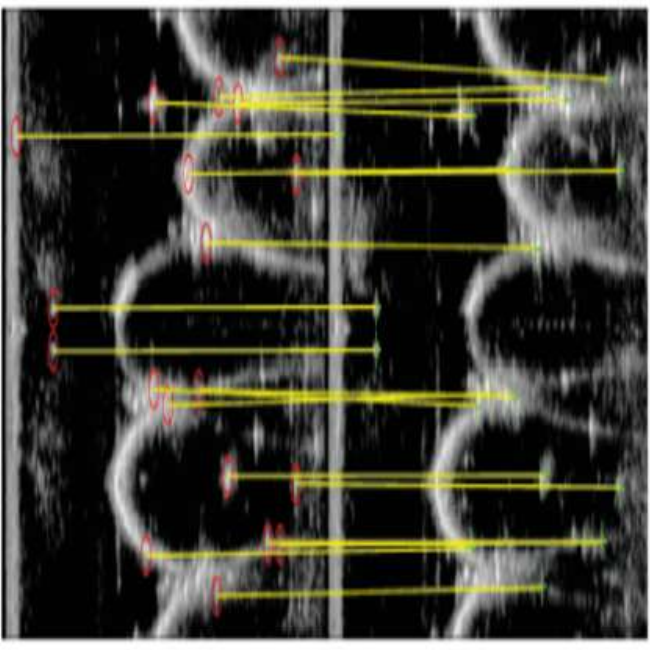} & \includegraphics[width=0.23\textwidth]{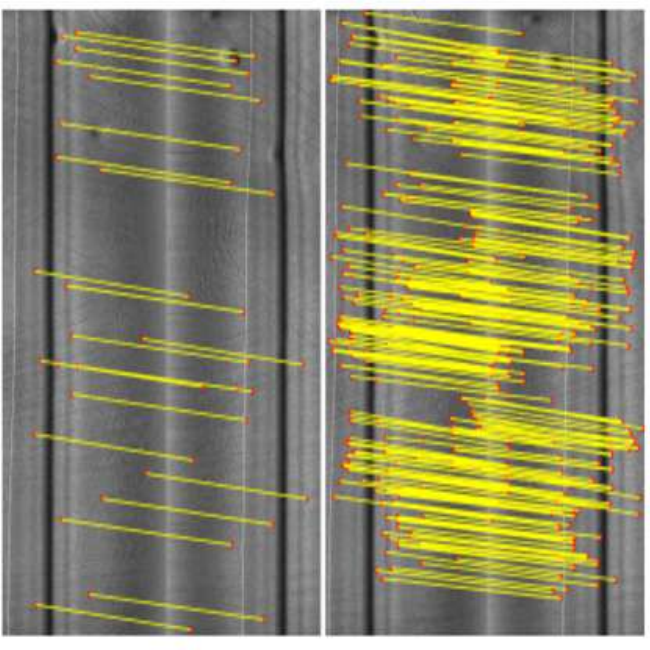}\\
    \end{tabular}}
    \caption{Samples SLAM on sonar images. Those two samples show a SLAM detected keypoint correspondence prediction on Mechanical Scanning Imaging Sonar (MSIS) and SSS images. Those points refer to identifying specific points of interest (key points) across multiple observations or images considered at the same physical location in the environment.}
    \label{sampleSLAMSonar}
\end{figure}
Deep SLAM \cite{b174} is an emerging research area that combines DL models with traditional SLAM. DL models enhance feature extraction, depth estimation, and image alignment in SLAM systems \cite{b21}, \cite{b22}, \cite{b23}, \cite{b24}. However, implementing SLAM in underwater environments poses several challenges, such as limited visibility (due to turbidity or darkness), underwater currents (causing AUVs to drift, leading to errors in localization and mapping), sensor noise (which affects pose estimation accuracy), and computational complexity (due to limited power and processing capacity of AUVs). Additionally, the lack of GPS underwater requires reliance on Kalman Filters (KF) \cite{kalmanFilter} to estimate vehicle position based on the last GPS fix and subsequent movements. Despite using KF or Extended Kalman Filters (EKF), the ground truth is still an "estimated" ground truth \cite{zhang2023dense}.  Recent methods combine MBES and optical data to leverage the strengths of both sensors while mitigating their limitations \cite{Jeremy-Slam}. A detailed overview of the challenges and technologies for underwater SLAM using sonar and optical sensors can be found in \cite{review-slam}. However, sonar-specific DL SLAM remains underdeveloped, presenting opportunities for future research, such as using a CNN-based model for feature extraction by W. Yang et al. \cite{CNN-Slam}.
\cref{sampleSLAMSonar} showcases two sonar samples where non-DL SLAM algorithms are applied. On the left, an EKF SLAM algorithm using Speeded-Up Robust Features (SURF) for feature extraction is applied to Mechanical Scanning Imaging Sonar (MSIS) images \cite{9705938}. On the right, Graph SLAM uses the Scale Invariant Feature Transform (SIFT) for feature extraction in SSS images \cite{Zhang_2023}.

\subsection{Challenges}
 
The primary concern when applying DL for sonar vision is the availability and quality of data for training the DL models. Regrettably, many companies and laboratories opt to keep their datasets private due to the high costs associated with data collection. Furthermore, there is a lack of comparative studies on sonar datasets in the literature, which could help practitioners train their models more effectively. On occasions when open-source datasets are available, they often do not match the specific objects, environmental conditions, sensors, or noise characteristics, resulting in an unusable dataset.
Consequently, many underwater robotics researchers must collect and annotate their data, which is time-consuming and costly. The quality of the sonar data is paramount for practical training, such as interference from other sonar devices, marine life, or general ambient noise, which can distort sonar data \cite{b27}. Thus, through this paper, we encourage researchers and practitioners to compare new models across datasets with varying characteristics when publishing to help readers understand the model's performance under different sonar conditions and environments, as demonstrated in \cite{UATD}. To help researchers and practitioners \autoref{Dataset - SOA} aims to present in a single document the SOA datasets for sonar-based tasks, such as classification, object detection, and segmentation collected with SSS and FLS. 
Although the lack of a dataset is the main critical point for applying reliable sonar-based DL for underwater tasks, the difference between training and deployment data is another important one. Indeed, as previously explained, sonar sensors suffer from different potential noise sources, which can differ drastically from the images in training. 
This discrepancy can lead to DL model errors, such as misclassification or failure to detect objects. Much research focuses on denoising sonar images to mitigate this issue, aiming to preprocess data to remove noise \cite{rs11040396}, \cite{9931211}, \cite{9790164}. While denoising is critical in sonar-based applications due to inherent noise issues, it faces three primary limitations. Firstly, the unpredictability of underwater noise may lead to scenarios where the denoising process is ineffective, leaving residual noise that can still fool DL models. This limitation underscores the challenge of ensuring comprehensive noise removal without sacrificing crucial data features. Another limitation is the potential loss of critical information during the denoising process. While denoising aims to improve image clarity by removing noise, it can inadvertently eliminate important features for accurately detecting and classifying underwater objects. Finally, recent denoising approaches for sonar images rely on autoencoders \cite{Liu2023} and deep autoencoder \cite{app14041341}, which, to be effective in real-time applications, must operate swiftly to avoid delaying the vehicle's interaction with potential objects. This requirement for speed can compromise the denoising quality or the overall system's efficiency, posing a trade-off between noise removal and operational effectiveness. Thus, even though denoising data increases the clarity of sonar images, more is needed to ensure the correctness of DL model prediction. Other alternative approaches, such as neural network verification, adversarial attack defenses, and out-of-distribution (OOD) detection, need further study to enhance the robustness of DL models in sonar applications beyond denoising techniques. Thus, the four last sections of \autoref{Safe AI - Sonar-Based Computational Vision DL} aim to provide the first robustness study under the scope of sonar neural network verification, adversarial attack, and OOD instead of denoising data.


\section{Sonar-Based Deep Learning: Toward Robustness}
\label{Safe AI - Sonar-Based Computational Vision DL}
This section aims to comprehensively summarize available open-source sonar datasets mostly for FLS and SSS sensors. It also covers the synthesis of simulators for generating simulated sonar data and explores various methods for generating synthetic data. Additionally, this section addresses the robustness of sonar-based DL models, detailing approaches like neural network verification, adversarial attacks, out-of-distribution detection, and uncertainty quantification. Conclusively, we propose a structured framework designed to enhance the robustness of sonar-based DL models, ensuring their reliability before deployment for real-world applications.

\subsection{Dataset - State of the art}
\label{Dataset - SOA}

In underwater exploration and research, sonar datasets are indispensable, offering a broad spectrum of applications from geological surveys to object detection. The previous section spotlighted the lack of open-source sonar datasets. Thus, this section is a detailed narrative synthesis of several notable datasets, highlighting their characteristics, utilities, and the nuances of their collection methods, aiming to provide better knowledge for researchers of open-source sonar datasets:
The UCI ML Repository—Connectionist Bench (Sonar, Mines vs. Rocks) Dataset \cite{misc_connectionist_bench} specializes in sonar signal intensity data for classifying mines and rocks. This dataset's strength lies in its focus on sonar signal characteristics, offering a resource for models designed to operate in environments where optical clarity is compromised. However, its specific focus on mines and rocks limit its direct application to broader image recognition or segmentation tasks.
S. Sugiyama et al. \cite{sugiyama2019sidescan} published SSS data from ice terraces in Glacier Grey, Patagonia \cite{LagoPaper} illuminating glacier formations. While offering ecological insights, this dataset's utility could be more constrained by the need for specialized software to interpret the data.
Ireland's Open Data Portal presents an eclectic collection of 14 sonar datasets \cite{irelandgov_sonar_datasets}, whose variety spans a broad spectrum of potential applications. Despite this diversity, the lack of detailed descriptions regarding the inclusion of SSS images or the nature of the datasets poses a challenge in identifying their applicability to specific research questions or projects.
The U.S. Government's Open Data portal \cite{USA_Dataset}, with its range of datasets, including sonar data, promises a wealth of data for various applications. However, the expansive scope of the portal makes locating specific types of sonar data, such as SSS images, a daunting endeavor that demands considerable time and effort.
The Marine\_PULSE dataset \cite{marine_pulse}, introduced by Du et al. \cite{MarinePulsePaper}, features side-scan sonar images focusing on underwater objects such as pipelines, cables, and engineering platforms. While the dataset's grayscale, object-focused images provide specificity, they are limited by low resolution and a lack of broader contextual features, which could impede comprehensive model learning.
\begin{figure*}[!t]
    \centering
    \begin{tabular}{cccc}
    Andreoli et al. \cite{Adriatic_Reefs} & Advaith et al. \cite{sethuraman2024machine} & Dahn et al. \cite{UXO} & Xing et al. \cite{marine_pulse}\\ 
    
    \includegraphics[width=0.20\textwidth]{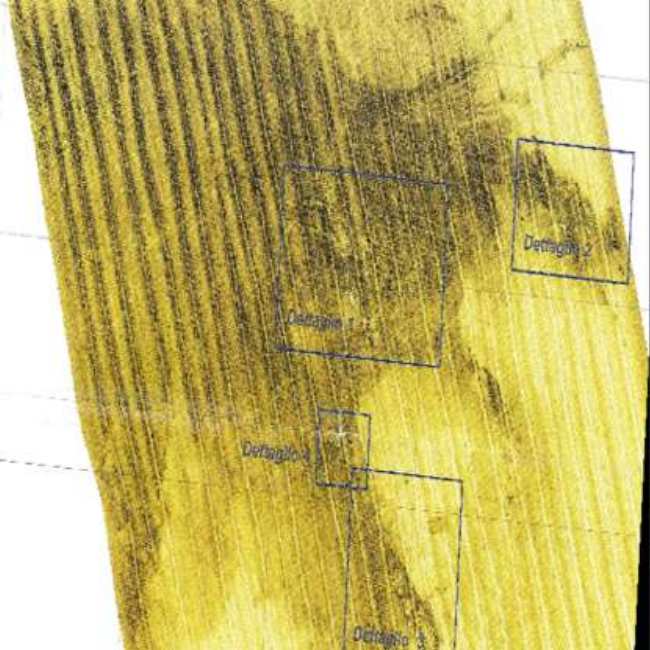} & \includegraphics[width=0.20\textwidth]{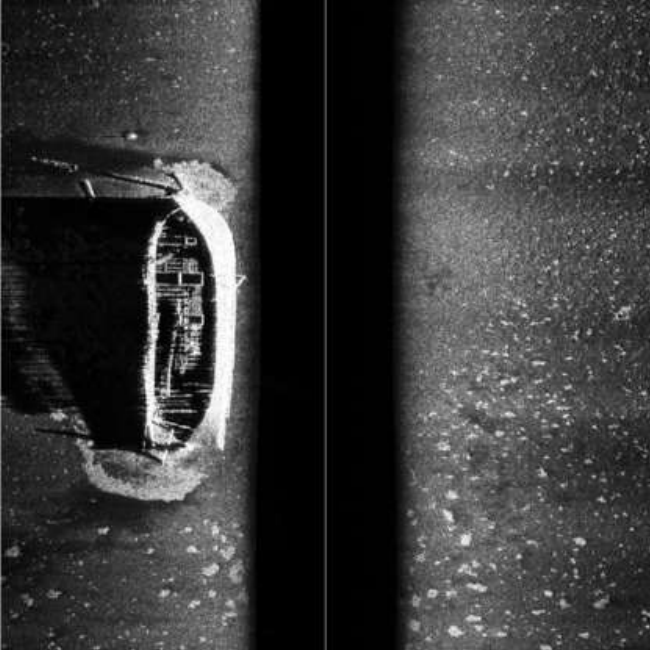}& \includegraphics[width=0.20\textwidth]{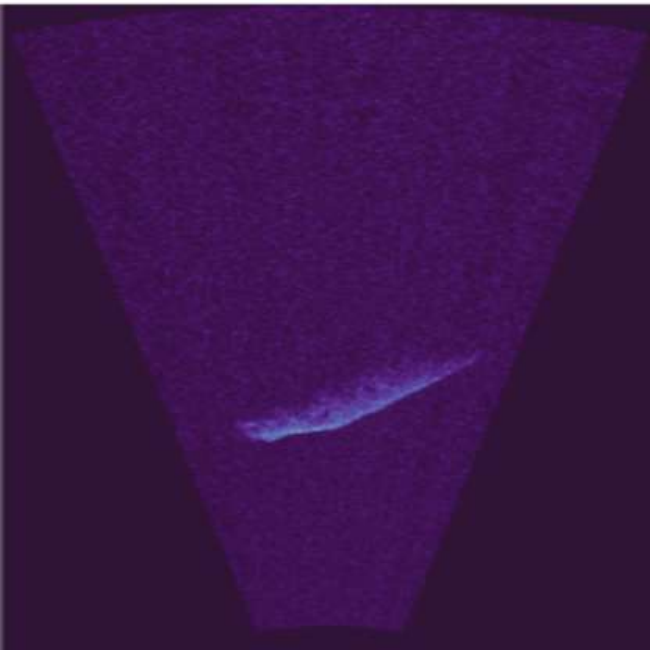}& \includegraphics[width=0.20\textwidth]{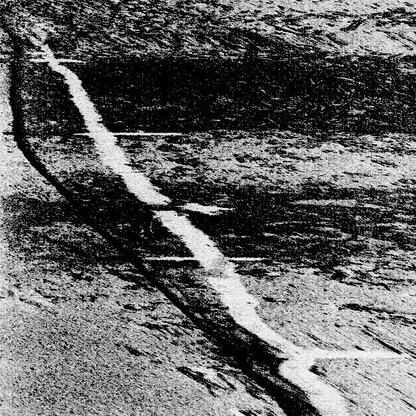}\\
    \end{tabular}
    \begin{tabular}{cccc}
    Martinez-Clavel et al. \cite{martinez_clavel} & Santos et al. \cite{Pessanha_Santos2024} & Álvarez-Tuñón et al. \cite{alvareztunon2024subpipe} & Aubard et al. \cite{aubard_2024_10528135}\\ 
    
    \includegraphics[width=0.20\textwidth]{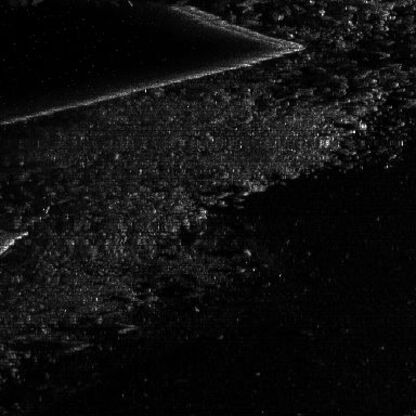}& \includegraphics[width=0.20\textwidth]{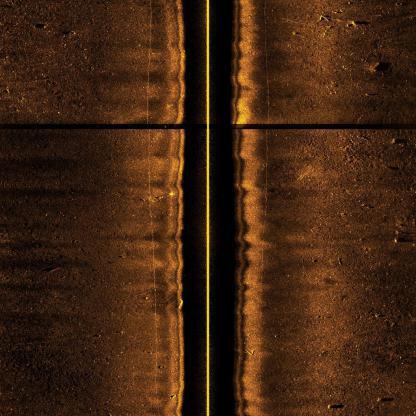}& \includegraphics[width=0.20\textwidth]{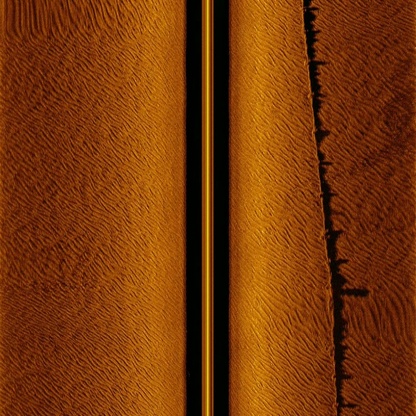}& \includegraphics[width=0.20\textwidth]{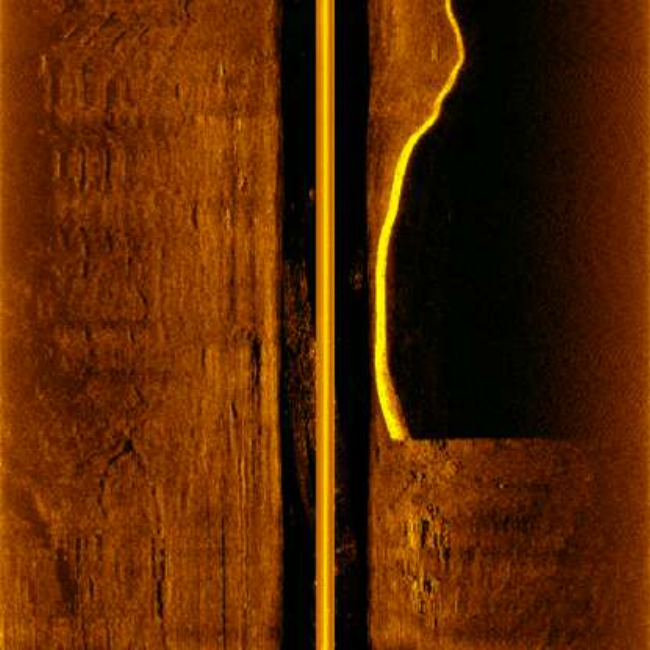}\\
    \end{tabular}
    \caption{Samples of Underwater Sonar Datasets. Those samples illustrate some of the SOA sonar datasets, which represent GeoTiff \cite{Adriatic_Reefs}, SSS \cite{sethuraman2024machine}, FLS \cite{UXO} images with different objects such as pipelines \cite{alvareztunon2024subpipe}, walls \cite{aubard_2024_10528135} and shipwrecks \cite{sethuraman2024machine}.}
    \label{fig:sample_SSS_dataset}
\end{figure*}
The Northern Adriatic Reefs dataset \cite{Adriatic_Reefs} offers georeferenced side-scan sonar mosaics of biogenic reefs off Venice. Although these low-resolution, whole-mosaic images capture broader geological formations, their format may need to be more conducive to training models to detect specific objects or features within a diverse underwater environment.
The Seafloor Sediments dataset \cite{martinez_clavel} boasts over 434164 images derived from side-scan sonar waterfalls, showcasing a variety of seafloor types along the coast of Catalunya. The dataset's large scale and seafloor-type diversity are invaluable. However, the lack of detailed information on the specific conditions of image collection (e.g., sonar range, altitude) might influence the generalizability of models trained on this data.
The UXO dataset \cite{UXO} focuses on unexploded ordnance detection in underwater environments, containing 74437 frames collected using the ARIS Explorer 3000 sonar in a controlled pool environment, with 48462 corresponding GoPro optical frames. Annotations, including bounding boxes and object types, are available for camera frames, supporting object detection tasks. Similarly, Side-scan Sonar Imaging for Mine Detection \cite{Pessanha_Santos2024}, with 1170 annotated images for distinguishing mine-like and non-mine-like objects, directly facilitates training in underwater mine detection. Despite its utility, the dataset would benefit from additional details on the original image sizes, sonar range, and AUV altitude during data collection, which is essential for understanding the detected object scale and appearance.
Despite its data augmentation potential, the focus on FLS images and the collection in a pool environment may only partially capture the complexities and variations of oceanic environments. The Underwater Acoustic Target Detection (UATD) dataset \cite{UATD} is an FLS dataset for object detection, which provides 14639 annotations among 9200 images. It provides a complete dual-frequencies dataset (720kHz, 1200kHz) with 10 classes: cubes, balls, cylinders, human bodies, planes, circle cages, square cages, metal buckets, tires, and bluerovs. The data collection setup is described in detail in \cite{UATD}, benefiting potential users using the dataset under the same conditions. Dual-Frequency Identification Sonar (DIDSON) dataset \cite{Ogburn2023} provides a fish dataset collected in the Rhode River, MD, USA, Indian River Lagoon, FL, USA, San Fransisco Bay, CA, USA, and Bocas del Toro, Panama, resulting in 100h of data where 1000 frames were extracted with eight fishes species labeled for segmentation model showcased by T. Perivoliot et al. \cite{DIDSON2}. The dataset requires the DIDSON-V5 software provided by Sound Metrics to convert the DDF format files. The Marine Debris Turntable (MDT) dataset \cite{MDT-Dataset} presented by D. Singh et al. \cite{singh2021marine} contains 2471 FLS images with 12 classes of objects, including bottles, pipes, platforms, and propellers annotated for the segmentation task. The Synthetic Aperture Sonar Seabed Environment Dataset (SASSED) \cite{SASSED} provides 129 complex-valued, high-frequency sonar snippets depicting various seafloor textures such as hardpack sand, mud, sea grass, rock, and sand ripple. Each snippet includes a hand-segmented mask image, which groups similar textures without necessarily representing ground truth labels, providing a valuable resource for training and testing sonar-based machine learning models.

The AI4Shipwreck dataset \cite{sethuraman2024machine} contributes to underwater archaeology and research. This dataset comprises 286 high-resolution SSS images collected from 24 distinct shipwreck sites within the Thunder Bay National Marine Sanctuary (TBNS), utilizing the EdgeTech 2205 SSS technology. A notable aspect of this collection is its focus on shipwrecks, offering a lens through which the underwater past can be explored and studied. Each image within the dataset has been annotated for segmentation tasks. 
The NKSID dataset \cite{NKSID-Dataset}, introduced by W. Jiao et al. \cite{JIAO2024123495}, provides a total of 2617 FLS images split into eight classes among big propellers, cylinders, fishing nets, floats, iron pipelines, small propellers, soft pipelines, tires, which makes it the most significant dataset for classification task among our dataset comparison. The data collection occurred in Bohai Bay with an ROV set-up with dual frequency (550kHz, 1.2MHz) Oculus M750d as FLS.
The SubPipe dataset \cite{alvareztunon2024subpipe} encompasses a comprehensive underwater collection, uniquely combining grayscale and RGB camera imagery with SSS images to offer a holistic view of underwater pipeline environments. Alongside vision (optical and sonar) data, this dataset enriches its offering with Conductivity, Temperature, and Depth (CTD) readings and navigational information, providing a multifaceted underwater exploration and analysis approach. Utilizing the Klein3000 sonar system at dual frequencies of 455kHz and 900kHz, the dataset captures high-quality imagery conducive to detailed study and model training. Annotations and benchmarks within the SubPipe dataset are tailored to evaluate SOA models across various applications, including segmentation, SLAM utilizing RGB data, and object detection using SSS images. The SSS dataset section of the Subpipe dataset counts 10030 images, making it the biggest open-source SSS object detection dataset in our comparison.  
SWDD (Sonar Wall Detection Dataset) \cite{aubard_2024_10528135} comprises 864 SSS images of walls, meticulously annotated following the COCO format, providing a resource for training and testing object detection models. The dataset includes an SSS waterfall video spanning 6 minutes and 57 seconds, from which 6243 images have been extracted and annotated to support benchmark validation efforts further. Including YOLOX \cite{b101} and YOLOX-ViT \cite{yolox-vit} models in the dataset's benchmarking process highlights the exploration of advanced object detection techniques in the context of SSS data. Utilizing the Klein3000 sonar at frequencies of 455kHz and 900kHz, the dataset offers high-resolution imagery conducive to detailed object detection tasks. It serves as a testing ground for innovative model architectures like YOLOX-ViT. In addition, an extended version of the dataset has been recently introduced through \cite{ROSAR}, where three datasets were collected under different weather conditions to improve object detection model comparison. These additional datasets include a total of 797 SSS images, which complement the original dataset. The Aurora dataset \cite{aurora} provides a multi-sensor collection for underwater exploration, integrating sonar, camera, and inertial sensors to facilitate the development of SLAM algorithms across different underwater environments. This comprehensive multi-sensor approach enriches SLAM research by enabling fusion between different types of sensory data, helping address the complexity of underwater localization and mapping. Krasnosky et al. \cite{krasnosky} presented a dataset that offers bathymetric surveys using MBES, complemented by high-precision GPS-based ground truth, which is particularly valuable for enhancing SLAM and bathymetric mapping methods. Similarly, Mallios et al. \cite{mallios} capture sonar imaging of underwater caves, a unique and highly challenging environment for SLAM testing. Those SLAM datasets highlight the importance of comprehensive sensory data and ground-truth validation to advance SLAM capabilities in diverse and demanding underwater environments.
\cref{fig:sample_SSS_dataset} represents some samples of those datasets. Each dataset's detailed account of data collection methods, including sonar range and environmental conditions, enriches underwater sonar research. Their diverse focuses—from geological formations to object detection—underscore the critical role of detailed documentation and the necessity for a broad range of datasets to address the multifaceted challenges of underwater exploration and monitoring.
\\
\begin{table*}[!htbp]
\caption{Comparison of the open source state-of-the-art sonar underwater datasets. This table compares the state-of-the-art sonar underwater dataset by analyzing the type of sonar (\textbf{Sonar}), type of data (\textbf{Data}), number of data (\textbf{No Data}), objects labeled in the data (\textbf{Object labels}), if the data is annotated, for which DL tasks (\textbf{Annotation}), if the data collection set up such as sonar frequency, altitude, etc. are described in the dataset or not (\textbf{Set-up}) and finally the year of the dataset publication (\textbf{Year}). The \textbf{*} is to mention that the dataset is not only limited to sonar but extended to other sensors such as optical cameras.} 
\centering
\footnotesize
\label{table:comparisonsoadataset}
\begin{tabular}{l@{\hspace{.3mm}} c@{\hspace{.8mm}} c@{\hspace{.8mm}} c@{\hspace{.4mm}} c@{\hspace{.3mm}} c@{\hspace{.3mm}} c@{\hspace{.3mm}} c@{\hspace{.3mm}}}
\hline
\textbf{ Dataset } & \textbf{ Sonar } & \textbf{ Data } & \textbf{ No Data } & \textbf{ Object labels } & \textbf{ Annotation } & \textbf{ Set-up } & \textbf{ Year } \\    
\hline
Northern Adriatic Reefs \cite{Adriatic_Reefs} & SSS & GeoTIFF & 7 & Reefs & \xmark & \xmark & 2010 \\
Lago Grey \cite{sugiyama2019sidescan} & SSS & Raw & \xmark & Glacier, Walls & \xmark & \cmark & 2019 \\

UCI ML \cite{misc_connectionist_bench} & \xmark & Raw & 211 & Mines, Rocks & Classification & \xmark & \xmark \\
SeabedObjects-KLSG \cite{SeabedObjects-KLSG} & SSS & Images & 1190 & Wrecks, Humans, Mines & Classification & \xmark & 2020 \\
Marine\_PULSE \cite{marine_pulse} & SSS & Images & 627 & Pipes, Mounds, Platforms  & Classification & \xmark & 2023 \\
NKSID \cite{NKSID-Dataset} & FLS & Images & 2617 & Infrastructures, Propellers, Tires & Classification & \cmark & 2024 \\

UATD \cite{UATD} & FLS & Images & 9200 & Tires, Mannequins, Boxes & Object Detection & \cmark & 2022 \\ 
SSS for Mine Detection \cite{Pessanha_Santos2024} & SSS & Images & 1170 & Mines & Object Detection & \xmark & 2024 \\
SWDD \cite{aubard_2024_10528135} & SSS & Images & 7904 & Walls & Object Detection & \cmark & 2024 \\
SubPipe \cite{alvareztunon2024subpipe} & SSS \textbf{*} & Images & 10030 & Pipelines & Object Detection & \cmark & 2024 \\
UXO \cite{UXO} & FLS & Images/Raw & 74437 & Unexploded Ordnances & Object Detection & \cmark & 2024 \\

MDT \cite{MDT-Dataset} & FLS & Images & 2471 & Infrastructures, Debris & Segmentation & \cmark & 2021 \\
SASSED \cite{SASSED} & SAS & Images & 129 & Muds, Sea Grass, Rocks, Sands & Segmentation & \xmark & 2022 \\
Seafloor Sediments \cite{martinez_clavel} & SSS & Images & 434164 & Rocks, Marine life & Segmentation & \cmark & 2023 \\
DIDSON \cite{Ogburn2023} & FLS & Images & 1000 & Fishes Species & Segmentation & \cmark & 2023 \\
AI4Shipwreck \cite{sethuraman2024machine} & SSS & Images & 286 & Shipwrecks & Segmentation & \cmark & 2024 \\

Cave Sonar \cite{mallios} & MSIS \textbf{*} & Rosbag & 500 meters & Cave Seabed & SLAM & \cmark & 2017 \\
Aurora \cite{aurora} & MBES, SSS \textbf{*} & Raw & 35h & Seabed, Marine habitats & SLAM & \cmark & 2020 \\
MBES-Slam \cite{krasnosky} & MBES & Rosbag & 4 missions & Seabed & SLAM & \cmark & 2022 \\
\hline
\end{tabular}
\end{table*}
\begin{figure}[!t]
    \centering
    {\begin{tabular}
    {cc}
    Training Data & Validation Data\\ 
    
    \includegraphics[width=0.22\textwidth]{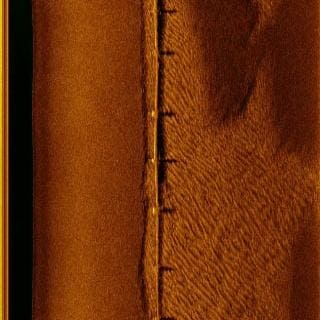} & \includegraphics[width=0.22\textwidth]{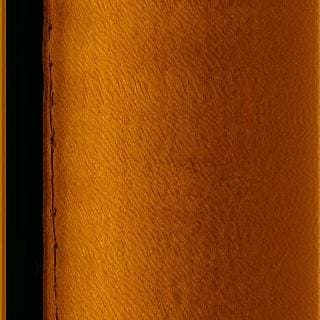}\\
    \end{tabular}}
    \caption{Comparison between training and validation datasets \cite{alvareztunon2024subpipe}. Those two samples present the difference between two SSS images from the exact location but were collected on two dates and at different altitudes. This comparison aims to explain that even using the same SSS at the same frequency and location, the SSS image can be drastically different, reducing the model's accuracy. Thus, this highlights that several parameters, such as AUV's altitude, need to be considered when using a DL model training dataset.}
    \label{fig:sample_SSS_difference}
\end{figure}

With the advent of DL on sonar data, a significant demand exists for datasets. Unlike unmanned aerial vehicles (UAVs) or autonomous ground vehicles, AUVs face unique data collection challenges due to the complexities of the underwater environment. This dataset comparison aims to highlight current sonar-based datasets. Traditionally, open-source sonar datasets for DL applications were rare, hindering the ability to compare different models effectively due to the utilization of disparate datasets. However, recently, a promising trend towards the publication and availability of such datasets, from Aubard et al. (2024) \cite{aubard_2024_10528135}, Álvarez Tuñón et al. (2024) \cite{alvareztunon2024subpipe}, Sethuraman et al. (2024) \cite{sethuraman2024machine}, W. Jiao et al. (2024) \cite{NKSID-Dataset}, N. Dahn et al. (2024) \cite{UXO} and Pessanha Santos et al. (2024) \cite{Pessanha_Santos2024} emerging in early 2024, alongside Martinez-Clavel et al. (2023) \cite{martinez_clavel}, M. Ogburn et al. (2023) \cite{Ogburn2023}, and Xing et al. (2023) \cite{marine_pulse} from 2023. This growing repository, although still modest in volume, signals a shifting paradigm towards enhancing the reproducibility of experiments and incentivizing collaborative data collection efforts within the research community.
Despite these advancements, sonar technology's inherent characteristics still require a delicate approach from training to inference data. Álvarez Tuñón et al. \cite{alvareztunon2024subpipe} illustrated how model performance may vary between training and operating datasets. They compare datasets collected at different dates and vehicle altitudes, resulting in 98\% in the average precision $AP_{50-90}$ for training, whereas only 15\% of $AP_{50-90}$ for inference data. \cref{fig:sample_SSS_difference} represents two samples from the training and validation dataset, where the pipeline size and shadow are different on both images despite the same environment. This variability underscores the importance of considering environmental and operational factors for training and operating.
\autoref{table:comparisonsoadataset} compares the SOA to the sonar datasets described. To the authors' knowledge, this is the most complete sonar dataset comparison. It presents a comprehensive overview of sonar datasets, including 10 SSS and 6 FLS datasets, encompassing 4 for classification, 5 for detection, 5 for segmentation, and 3 for SLAM, alongside two datasets lacking annotations. It reveals that the NKSID \cite{NKSID-Dataset} provides the most images in classification with 2617 images, while SubPipe \cite{alvareztunon2024subpipe} and UXO \cite{UXO} in detection respectively with 10030 and 74437 images. In contrast, UATD \cite{UATD} offers a broader scope with 10 distinct classes across 9200 images. The most complete dataset for segmentation is Seafloor Sediments \cite{martinez_clavel}, boasting 434164 images. 

Moreover, unlike terrestrial and aerial optical images, which can be benchmarked with established datasets like COCO \cite{coco_dataset}, or ImageNet \cite{imagenet_dataset} (for object detection), sonar datasets have yet to establish a universally recognized baseline dataset. This absence complicates direct model comparisons, highlighting a current limitation for future development within the community. Establishing such a benchmark would greatly facilitate advancements in sonar-based DL, improving the field toward greater standardization and comparability.
Thus, due to this limitation, we have created a dedicated GitHub repository centralizing current open-source underwater sonar datasets, providing a foundation for future benchmarking efforts. In addition, researchers and practitioners are encouraged to contribute by adding new datasets, ensuring the resource remains comprehensive and up to date. The repository is available at \url{https://github.com/remaro-network/OpenSonarDatasets}.
 

\subsection{Synthetic data}

\begin{table*}[!htbp]
\caption{Comparative Analysis of the Open-source underwater simulators. This table compares the underwater simulator and provides insight into the vision sensors those simulators provide. The table analyzes the date of publication (\textbf{Year}), the ROS support (\textbf{ROS Support}), then if they provide vision sensors among Camera (\textbf{Camera}), underwater LIDAR (\textbf{LIDAR}), Forward Looking Sonar (\textbf{FLS}), Side Scan Sonar (\textbf{SSS}), and Mechanical Scanning Imaging Sonar (\textbf{MSIS}).}
\centering
\footnotesize
\label{table:comparisonsSimulators}
\begin{tabular}{l@{\hspace{.3mm}} c@{\hspace{.8mm}} c@{\hspace{.8mm}} c@{\hspace{.4mm}}  c@{\hspace{.3mm}} c@{\hspace{.3mm}} c@{\hspace{.3mm}} c@{\hspace{.3mm}}}
\hline
\textbf{Simulator} & \textbf{Year} & \textbf{ ROS Support } & \multicolumn{5}{c}{\textbf{Vision Sensors}} \\

& & & \textbf{Camera } & \textbf{ Lidar } & \textbf{ FLS } & \textbf{ SSS } & \textbf{ MSIS }\\
\hline
UWSim \cite{UWSim}  & 2012 & ROS 1 & \cmark & \cmark & \xmark & \cmark \cite{UWSimSSS} & \xmark \\
UUV \cite{b192} & 2016 & ROS 1 & \cmark & \cmark & \xmark & \xmark & \xmark \\
ImagingSonarSimulator \cite{CERQUEIRA2017} & 2017 & ROS 1 & \xmark & \xmark & \cmark & \xmark & \cmark \\
DAVE \cite{DAVE} & 2022 & ROS 1 & \cmark & \cmark & \cmark & \xmark & \xmark \\
Stonefish \cite{b193} & 2019 & ROS 1 \& 2 & \cmark & \xmark  & \cmark & \cmark & \xmark \\
UNavSim \cite{b201} & 2023 & ROS 1 \& 2 & \cmark & \cmark & \xmark & \xmark & \xmark \\
MARUS \cite{MARUS} & 2022 & ROS 1 \& 2 & \cmark & \cmark & \cmark & \xmark & \xmark \\
HoloOcean \cite{Holocean} & 2022 & ROS 2 & \cmark & \xmark & \cmark & \cmark \cite{HoloceanSonar} & \xmark \\
\hline
\end{tabular}
\end{table*}
Given the limited open-source sonar dataset, a novel research trend relies on simulated data for real-world implementations, known as Simulation-to-Real \cite{tobin2017domain}. The goal is to reduce the costly and time-consuming process of underwater data collection by relying on underwater simulators equipped with sensor payloads to generate simulated data, resulting in the development of multiple open-source simulators.
UWSim \cite{UWSim} is the first underwater simulator designed for marine robotics applications, built on the OpenSceneGraph and osgOcean libraries to provide realistic underwater vision and physics that support various sensors and vehicles. D. Gwon et al. \cite{UWSimSSS} improved the UWSim simulator by proposing an SSS plugin. DAVE \cite{DAVE} and UUV Simulator \cite{b192}, anchored in the ROS \cite{ROS} and Gazebo framework, primarily support conventional cameras and FLS to enhance acoustic visualization, making them ideal for users engaged with the ROS ecosystem. Furthermore, DAVE provides an underwater point cloud LIDAR sensor. Stonefish \cite{b193} uses ROS for publishing virtual sensor measurements and includes cameras, FLS, and SSS images, underscoring its proficiency in simulating custom acoustic data. However, the quality of simulated SSS images still needs to match actual SSS data. UNavSim \cite{b201}, compatible with ROS and leverages high-detail rendering Unreal Engine 5 and AirSim \cite{Airsim}, brings camera and underwater LIDAR technologies. MARUS \cite{MARUS}, also compatible with ROS, offers a comprehensive sensor suite including underwater LIDAR, cameras, and FLS, highlighting the simulator's commitment to providing diverse and realistic sensor data for underwater research. HoloOcean \cite{Holocean} also generates camera and FLS imagery to craft realistic underwater scenarios. Furthermore, E. Potokar et al. recently improved their HoloOcean simulator \cite{HoloceanSonar} by implementing multibeam imaging, multibeam profiling, SSS, and echo-sounder, which makes it one of the most complete simulators for simulated sonar images. R. Cerqueira et al. \cite{CERQUEIRA2017} present ImagingSonarSimulator, a sonar simulator for FLS and Mechanical Scanning Imaging Sonar (MSIS). Using the Rock-Gazebo framework \cite{7402154}, it models physical forces in the underwater environment, providing real-time simulation for a virtual AUV. The simulator uses the OpenGL shading language (GLSL) \cite{GLSL} on a GPU to emulate sonar devices based on parameters like pulse distance, echo intensity, and field-of-view. Furthermore, the simulator is compatible with ROS 1. The paper in \cite{simuAnalysis} provides a deeper review of sonar and non-sonar simulator analysis. This comparison underscores that while underwater simulators are progressively advancing as a research domain, offering increasingly lifelike representations of underwater environments through cameras and FLS, they still confront challenges in providing and accurately replicating the nuances of real SSS imagery. \autoref{table:comparisonsSimulators} systematizes the underwater simulators by highlighting their vision-based sensors, such as camera and underwater lidar, providing 3D point cloud, FLS, SSS, and MSIS. This comparison shows that most simulators focus primarily on rendering camera images, while HoloOcean \cite{HoloceanSonar} and Stonefish \cite{b193} provide an extensive range of realistic sonar capabilities. The GitHub repository MASTODON \cite{MASTODON-github} provides a novel sonar image simulator, as described by D. Woods \cite{MASTODON}. However, the paper is not open-access, limiting detailed information availability.
This section analysis focuses exclusively on open-source sonar simulators to support reproducible research and open-access studies. Nevertheless, non-open-source sonar simulators, such as E. Coiras et al. \cite{GPU-based-simulator}, are also available in the literature. 

However, despite reducing data collection time, simulated datasets usually represent ideal conditions, while the real world contains many uncertainties. For instance, sonar images may require adding noises to align them with real-world sonar data. This difference between simulated and real data is known as the sim-to-real gap, an ongoing research topic providing promising results for Object Detection \cite{b175} and Segmentation \cite{b195}. Generative Adversarial Networks (GANs) \cite{b196}, traditionally used in tasks like Natural Language Processing (NLP) \cite{b197} and image generation, have emerged as a promising avenue for underwater image generation. Current research explores the potential of GANs to merge simulated and real data, thereby producing expansive datasets \cite{b198}. N. Jaber et al. \cite{Sonar2Depth} use conditional GAN (cGANs) for increasing their FLS dataset collecting with the Stonefish simulator and validating it in a pool environment.
Similarly, E. Lee et al. \cite{data_aug_dataset} address the scarcity of sonar imagery by employing a Pix2Pix-based cGAN to generate synthetic sonar images. This method enhances segmentation model training by simulating diverse underwater conditions, leading to improved performance when real data is limited, highlighting that GANs could offer a viable solution to the prevailing challenge of insufficient underwater data. Diffusion models \cite{diffusion_model}, another emerging class of generative models, employ a sophisticated process of progressively adding then removing noise from images, offering stable training and diverse outputs, and showing improvement in SOA object detection on sonar images \cite{diffusion_model_sss}. In contrast, GANs create data through a competitive process between a generator and a discriminator, known for producing highly realistic images but with potential training instability. G. Huo et al. \cite{SeabedObjects-KLSG} propose a semisynthetic data generation method to generate data from optical to sonar data of airplanes and drowning victims using image segmentation with intensity distribution. This semisynthetic method aims to crop the object on optical images and add specific shadows, sonar backgrounds, and sonar distributions, resulting in an image that looks like a sonar image. Following the same principle, Z. Bai et al. \cite{10399359} propose a global context external-attention network (GCEANet), which, from optical images, produces pseudo-SSS images corresponding to the absent categories for zero-shot SSS image classification.
Furthermore, Data Augmentation and Transfer Learning, two well-known methods for improving DL outputs, can also be implemented during the DL training process to improve the accuracy and robustness of models. Data Augmentation \cite{b199} increases the size of the dataset by filtering, rotating, and adding random noises in the original dataset, improving the accuracy and model performance against noise. Transfer Learning enhances the model's accuracy \cite{b200}. Instead of starting the DL model training from scratch, pre-trained DL weights with bigger datasets are transferred into the DL model before training with the underwater dataset. 
\\

\subsection{Neural Network Verification}

Neural network verification \cite{NNV} in the context of sonar aims to ensure that DL models can accurately interpret data under various noise conditions within defined boundaries. It validates the model's predictions against expected outcomes across all potential sonar inputs, ensuring reliability in diverse underwater environments. Applying DL to real-world scenarios, especially in challenging underwater settings, requires robust verification to ascertain reliability and robustness \cite{huang2020survey}.
In image processing, DL models typically incorporate various layers—such as convolutional, pooling, and fully connected—and nonlinear activation functions like ReLU, softmax, and sigmoid. 
The challenge of verification is amplified by the high-dimensional input space and the complex nonlinearities introduced by activation functions.
Several tools have been developed to address this challenge. Alpha-Beta-CROWN \cite{zhang2021beta} uses optimized bounding methods to determine neuron activation bounds, efficiently reducing computational demands. ERAN \cite{ERAN} and DeepPoly \cite{DeepPoly} utilize abstract interpretation techniques to balance verification precision and scalability, while Reluplex \cite{reluplex} and Marabou \cite{marabou} focus on verifying ReLU-based networks, with Marabou extending support to more architectures and activation functions.
These tools are valuable for ensuring robustness in DL models used for image classification and object detection, where reliability is critical. However, despite the advancements in neural network verification, current tools are primarily designed for classification tasks and face limitations when applied to other areas such as object detection, segmentation, and SLAM. Additionally, adapting these methods to sonar data introduces new challenges due to the unique uncertainties of underwater environments, which require greater focus in future research.

\subsection{Adversarial Attack}

Neural network Verification methods often fail to keep pace with the growing complexity of SOA DL models, leading to significant demands on computational resources due to the nonlinearities and high dimensionality of DL models, making scalability a critical concern for researchers and practitioners \cite{CandN}, \cite{FGSM}. The Fast Gradient Sign Method (FGSM) \cite{FGSM} is the first white-box adversarial attack methodology that leverages the gradients of a neural network to craft adversarial examples efficiently and quickly. Building upon FGSM's foundation, Projected Gradient Descent (PGD) \cite{PGD} introduces an iterative approach that enhances the basic concept. PGD refines FGSM's technique by applying multiple small-step adjustments to the input, ensuring the perturbation remains within a defined epsilon neighborhood of the original input, enhancing the adversarial example's effectiveness while maintaining its subtlety. DeepFool \cite{Deepfool} diverges from PGD and instead focuses on the precision of perturbations; it seeks the minimal change required to alter a model's classification, providing a more nuanced estimation of model robustness. Carlini \& Wagner's (C\&W) Attack \cite{CandN} identifies the minor possible perturbation that can still mislead a model, comparable to DeepFool's principle of minimal disruption. However, C\&W designs their attack to be effective even against models fortified with defensive measures. In a targeted vector, the Jacobian-based Saliency Map Attack (JSMA) \cite{JSMA} focuses on strategically modifying input features. By exploiting the output gradient concerning the input, JSMA identifies critical pixels whose alteration would specifically misguide the model into a chosen misclassification.
As highlighted by N. Papernot et al. \cite{Papernot}, transfer attacks exploit the phenomenon of perturbation transferability, highlighting the ability of adversarial examples to mislead different models. Query-based strategies, explored by P. Chen et al. \cite{ZOO} involves iterative input adjustments based on model feedback, aiming to find adequate adversarial inputs without accessing the model's gradients. Decision-based attacks, such as those by Brendel et al. \cite{Brendel}, refine adversarial inputs using model outputs, score-based attacks, and Ilyas et al. \cite{Ilyas} using confidence scores. 

\begin{figure}[!t]
    \centering
    {\begin{tabular}
    {cc}
    SSS - Black Lines & SSS - Clean\\ 
    
    \includegraphics[width=0.23\textwidth]{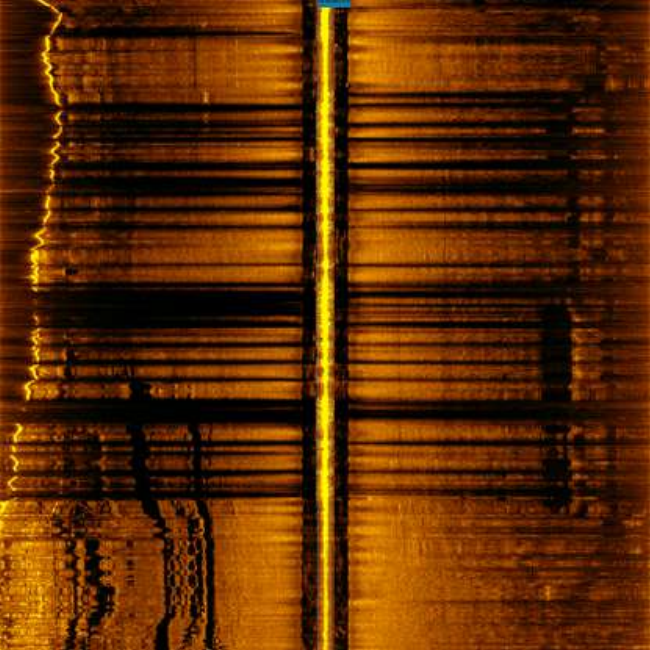} & \includegraphics[width=0.23\textwidth]{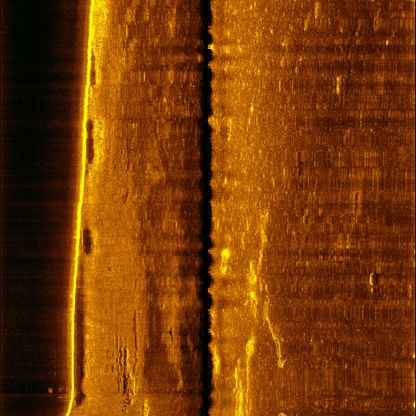}\\
    \end{tabular}}
    \caption{Comparison between Side Scan Sonar with and without loss of information from \cite{ROSAR}, characterized by black lines. The image on the left represents an SSS image with a loss of information, whereas the image on the right is without and with a nadir gap filtering. This loss of information appeared during a mission while the vehicle was at the surface, and the weather was windy, which resulted in the transducer going out of the water, represented by black lines on the image. This unexpected event resulted in wall non-detection and highlighted the lack of the model's robustness under noise.}
    \label{fig:SSS_noise}
\end{figure}

Most perception studies focus on grounded or aerial datasets for UAVs \cite{b57}, predominantly working with RGB images \cite{b58}. However, underwater robots that rely mostly on sonar vision require a study of their specific uncertainties, which differ from RGB cameras.
\cref{fig:SSS_noise} shows an SSS sample with and without signal loss from \cite{ROSAR}, characterized by black lines in the image. These samples were collected at the surface during a storm, causing the sonar transducers to emerge from the water, resulting in missing data occasionally. Although this information loss is not adversarial noise in the traditional sense, it is a natural adversarial scenario, highlighting the unpredictable environmental conditions that can affect model performance. Such scenarios can be leveraged in adversarial retraining to enhance model robustness.
Several laboratories are exploring adversarial methods for sonar images due to the growing interest in underwater tasks. ROSAR framework, proposed by Aubard et al. \cite{ROSAR}, utilizes PGD attacks to target specific safety properties of SSS, similar to the type of signal loss shown in \cref{fig:SSS_noise}. When these properties are compromised, the method generates counterexamples for adversarial retraining, resulting in up to 1.85\% in detection model robustness. Q. Ma et al. \cite{FasterRcnnAdversarialAttack} propose the Noise Adversarial Network (NAN), which introduces noise into the dataset and applies it to the Faster R-CNN object detection model, enhancing robustness by 8.9\% mAP on a sonar dataset. Furthermore, Q. Ma et al. \cite{b77} present the Lambertian Adversarial Sonar Attack (LASA), an adversarial attack for side-scan sonar images based on the Lambertian reflection model. They compare LASA with FGSM, PGD, and Deepfool and conclude that LASA significantly improves the robustness of sonar-based classifiers. S. Feng et al. \cite{DLadversarialAttack} examine the impact of adversarial noise on both CNNs and transformers when applied to sonar spectrograms.

\subsection{Out-Of-Distribution}
Sonar images suffer from the disparity between training data and real-world sonar inputs, variations in sonar brands, operating frequencies, and environmental conditions, which presents a substantial challenge known as Out-Of-Distribution (OOD) \cite{b52}. Hendricks and Gimpel \cite{hendrycks2017a} introduced softmax probabilities to distinguish between correctly classified, misclassified, and OOD examples within neural networks. Building on this, Liang et al. \cite{liang2018enhancing} enhanced the model's ability to differentiate in-distribution (ID) from OOD data through temperature scaling and input preprocessing, introduced by their ODIN technique. Lee et al. \cite{Lee2018ASU} proposed the Mahalanobis distance for a sophisticated similarity measure between input features and class-conditional distributions, improving the efficacy of OOD detection. Generative models such as GANs and Variational Autoencoders (VAEs) have enabled the identification of OOD inputs by learning latent representations of ID data. Kingma and Welling's \cite{kingma2022autoencoding} work, alongside Higgins et al.'s introduction of $\beta$-VAEs \cite{higgins2017betavae}, further enhance this by controlling differentiation in latent spaces, thereby improving OOD detection. 

In applying these advancements to sonar imaging, I. Gerg et al. \cite{OOD-SAS} directly addresses OOD detection by incorporating a Perceptual Metric Prior (PMP) within the training classification loss. This approach notably improves the robustness of models, especially for sonar image classification tasks characterized by limited data and subtle distribution differences. It presents an innovative alternative to conventional methods like data augmentation and hyperparameter tuning, specifically tailored to overcome the unique challenges posed by sonar data analysis. Furthermore, W. Jiao et al. \cite{Sonar-Long-Tail} introduces the Balanced Ensemble Transfer Learning (BETL) framework to address the compounded challenges of long-tail and few-shot classification in sonar images. This framework enhances classification accuracy while optimizing memory and inference time, indirectly aiding OOD detection by improving model performance on sparsely represented classes that often resemble OOD samples. As an extension of \cite{Sonar-Long-Tail}, W. Jiao et al. \cite{JIAO2024123495} present the first comprehensive examination of open-set long-tail recognition (OLTR) specifically for sonar images, marking an improvement in sonar-specific OOD detection. This work navigates into the difficulties of classifying sonar data and setting new benchmarks by evaluating SOA algorithms and proposing the novel PLUD (Push the Right Logit Up and the wrong Logit Down) loss function. M. Cook et al. use their NuSA (Null Space Analysis) approach \cite{NuSA}, which aims to detect outliers while testing for classification tasks, to detect unknown objects during automatic target recognition tasks in sonar data \cite{NuSA-Sonar}, and conclude that NuSA applies to sonar images outperform the OOD methods such as Unsupervised Self-Supervised Outlier Detection (SSD) \cite{SSD-OOD} and  Rectified Activation (ReAct) \cite{REACT-OOD}.

\subsection{Uncertainty Quantification}

Uncertainty quantification in DL models serves as a critical framework for assessing the confidence and reliability of predictions. It is categorized into two main types: aleatory uncertainty, coming from the inherent noise and variability in the data \cite{MacKay2003}, and epistemic uncertainty, which stems from the model's lack of knowledge or uncertainty about the model itself \cite{Rasmussen2004}. Techniques for addressing these uncertainties have been extensively reviewed, highlighting the dual nature of predictive uncertainty within supervised learning frameworks \cite{ABDAR2021243} \cite{H_llermeier_2021}. The methodologies range from Bayesian inference, which offers a deep-rooted framework for epistemic uncertainty, to heteroscedastic neural networks that effectively model aleatory uncertainty by allowing variance in predictions based on data noise \cite{kendall2017uncertainties}. Monte Carlo Dropout, introduced by Y. Gal et al. \cite{gal2016dropout}, presents a more computationally feasible approximation of Bayesian inference, balancing practicality and theoretical rigor. Addressing aleatory uncertainty, heteroscedastic neural networks, as explored by A. Kendall et al. \cite{kendall2017uncertainties}, propose a model adjusting its confidence levels based on the inherent noise present in input data. This approach models uncertainty as a function of the data, allowing predictions to reflect the variability in the underlying data distribution. Ensemble methods, introduced by B. Lakshminarayanan et al. \cite{lakshminarayanan2017simple}, emerge as a robust approach to encapsulate both types of uncertainty by aggregating predictions from a collection of models, which can capture model variance and reflect data variability. Despite their effectiveness, ensembles require multiple models to be trained and maintained, which may not be feasible in resource-constrained environments. When applied to sonar-based data, uncertainty quantification in DL models becomes crucial for enhancing the robustness and reliability of underwater object detection and classification tasks. L. Fuchs et al. \cite{9775246} propose a pipeline for generating simulated FLS data using cycleGAN and ensuring the quality of the simulated images for real deployment by analyzing the data uncertainty for detection and classification tasks. P. Tarling et al. \cite{Tarling2022} combined self-supervised learning with uncertainty quantification to improve training and measure prediction uncertainty for fish detection on FLS images. By estimating the noise variance of the dataset images, they adjust the loss function to regulate the aleatoric uncertainty.

\subsection{Limitations and Workflow}
\begin{table*}[!htbp]
\caption{Comparative Analysis of the previous works focused on sonar-based robustness separated into three main sections: OOD, Adversarial Attack, and Uncertainty Quantification. This table analyses the date of publication (\textbf{Year}), the robustness method (\textbf{Robustness}) where \textit{AA} stands for Adversarial Attack, \textit{OOD} for Out-Of-Distribution and \textit{UQ} for Uncertainty Quantification, the method used in their papers (\textbf{Method}), the type of sonar use (\textbf{Sonar}), and the DL task realized (\textbf{Task}) where \textit{C} stands for classification and \textit{D} for detection.}
\centering
\footnotesize
\label{table:comparisonsRobustness}
\begin{tabular}{l@{\hspace{.3mm}} c@{\hspace{.8mm}} c@{\hspace{.8mm}} c@{\hspace{.4mm}} c@{\hspace{.4mm}} c@{\hspace{.4mm}}}
\hline
\textbf{Paper} & \textbf{ Year } & \textbf{ Robustness } & \textbf{ Method } & \textbf{ Sonar } & \textbf{ Task }\\
\hline
 Q. Ma et al. \cite{FasterRcnnAdversarialAttack} & 2020 & AA & Noise Adversarial Network (NAN) & SSS and SAS & D \\
S. Feng et al. \cite{DLadversarialAttack} & 2023 & AA & FGSM \cite{FGSM} and PGD \cite{PGD} on Spectrogram & Spectrogram & C\\
M. Shell et al. \cite{b77} & 2021 & AA & Lambertian Adversarial Sonar Attack (LASA) & SSS & C\\
M. Aubard et al. \cite{ROSAR}  & 2024 & AA & ROSAR & SSS & D\\
I. Gerg et al. \cite{OOD-SAS} & 2023 & OOD & Perceptual Metric Prior (PMP) & SAS & C \\
W. Jiao et al. \cite{Sonar-Long-Tail} & 2022 & OOD & Balanced Ensemble Transfer Learning (BETL) & FLS & C \\
W. Jiao et al. \cite{JIAO2024123495} & 2024 & OOD & PLUD & FLS & C \\
M. Cook et al. \cite{NuSA-Sonar} & 2020 & OOD & NuSA \cite{NuSA} on Sonar & SAS & D  \\
L. Fuchs et al. \cite{9775246} & 2022 & UQ & cycleGAN & FLS & D, C\\
P. Tarling et al. \cite{Tarling2022} & 2022 & UQ & Prediction Uncertainty & FLS & D\\

\hline
\end{tabular}
\end{table*}

\Cref{Safe AI - Sonar-Based Computational Vision DL} aims to help researchers and practitioners interested in leveraging sonar-based DL perception tasks by comparing and reviewing state-of-the-art sonar datasets and simulators into a consolidated document, facilitating access to simulated and field-collected dataset. Surprisingly, our research revealed a lack of comprehensive surveys of these datasets, mainly because researchers were not used to revealing their datasets. However, this scenario is gradually changing with a novel trend toward data sharing, as evidenced by the recent availability of numerous open-source datasets from 2023 and 2024. This shift underscores a growing commitment within the research community towards openness and collaboration, significantly benefiting the field.
The section also highlights the unique sonar uncertainties encountered underwater, affecting DL models' performance. Discrepancies between training and operational data, including variations in environmental conditions, sonar setup, and seabed characteristics, can potentially mislead models. Presently, the focus within sonar methodologies leans heavily towards image denoising before model inference rather than supporting the intrinsic robustness of the models themselves. Consequently, this section presents strategies for addressing uncertainties inherent in sonar imaging and the first survey focusing on robustness for sonar-based DL models. However, this preliminary survey highlights a noticeable scarcity of research explicitly addressing neural network verification (0 papers), adversarial attacks (4 papers: \cite{FasterRcnnAdversarialAttack}, \cite{DLadversarialAttack}, \cite{b77}, \cite{ROSAR}), OOD detection (4 papers: \cite{OOD-SAS}, \cite{Sonar-Long-Tail}, \cite{JIAO2024123495}, \cite{NuSA-Sonar}), and uncertainty quantification (2 papers: \cite{9775246}, \cite{Tarling2022}), as detailed in \autoref{table:comparisonsRobustness}.
Through this paper, we encourage further research into the robustness of sonar-based DL models, which would enhance the safety and autonomy of underwater robotic systems.
\begin{figure}[!t] 
    \begin{center}
        \noindent \includegraphics[scale=0.45]{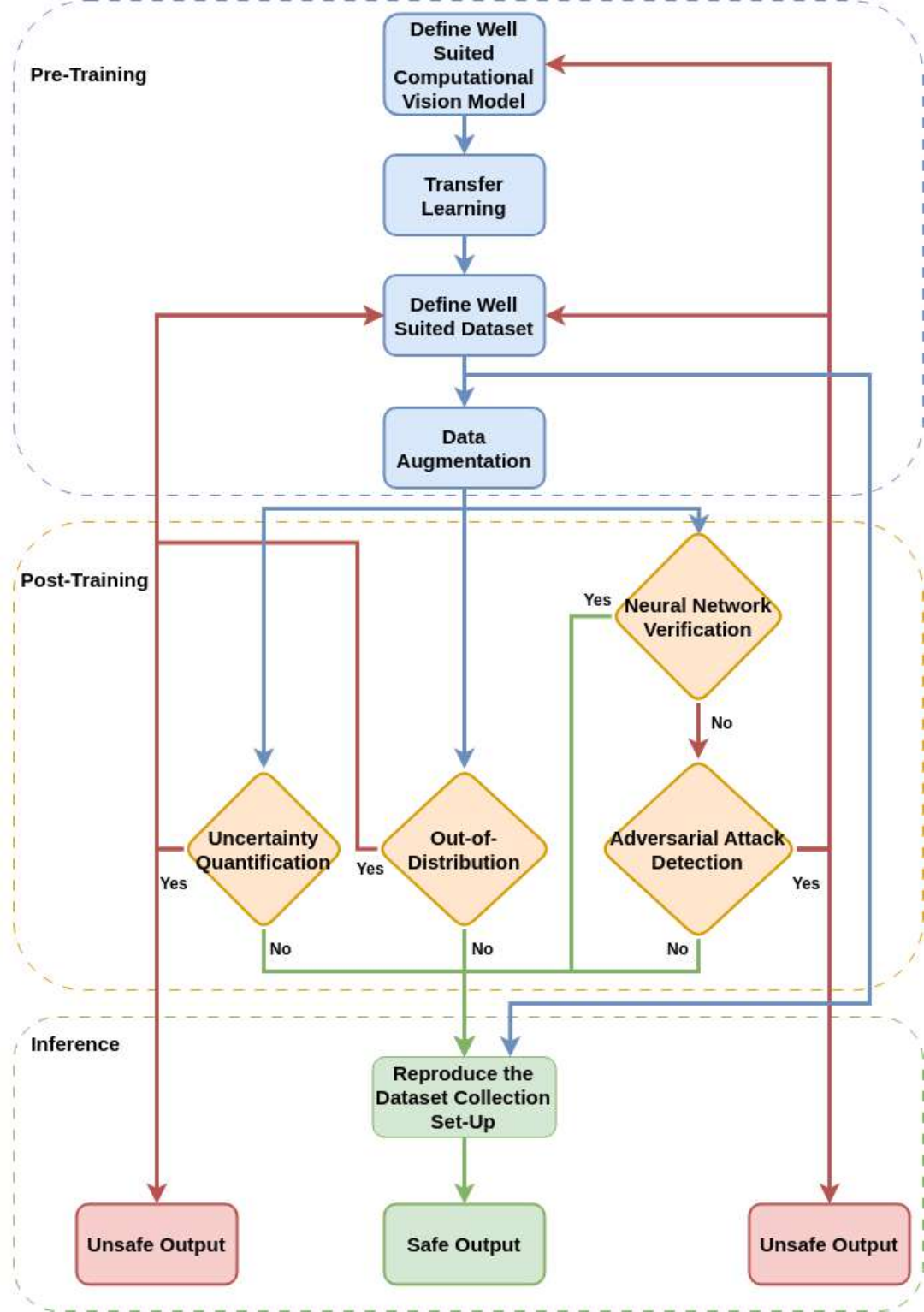}
                \caption{Sonar-based Deep Learning - Robustness Workflow. This proposed workflow introduces two main steps: the pre-training step, represented by blue boxes (define well-suited computer vision model, transfer learning, define well-suited dataset and data augmentation), and the post-training steps before deploying a sonar-based DL model, represented by orange diamond, which should result in better model prediction under unexpected noises (neural network verification and adversarial attack detection) and underwater characteristics (out-of-distribution, epistemic uncertainty). Finally, suppose the output of the OOD, uncertainty quantification (epistemic and aleatory), and neural network verification (or adversarial attack detection) do not return any error; the sonar set-up from the dataset collection should be reproduced (frequency, vehicle altitude) to ensure the correct model behavior during inference.}
                \label{Safe Vision AI}               
    \end{center}
\end{figure}

To ensure the robustness of sonar-based DL model implementation into AUVs, we support our document by proposing a framework describing the specific steps improving the models' robustness.
\cref{Safe Vision AI} describes a comprehensive workflow by selecting a well-suited computer vision model and dataset, applying transfer learning and data augmentation. If the unsafe output corresponds to OOD data or uncertainty quantification, the dataset requires refinement to enhance detection reliability. Furthermore, the model must undergo either a neural network validation or adversarial attacks under specific underwater noises. If those two steps are invalid, the model is considered 'unsafe,' requiring a revision of the model and training dataset.
'Safe output' signifies the model's ability to perceive its environment, resulting in potential correct vehicle behavior. However, the training dataset setup should match the inference setup, including sonar frequency, color map, and vehicle altitude.
In the context of 'unsafe,' decisions by the autonomous vehicle driven by the DL model could engender unwanted outcomes, resulting in hazardous scenarios.
For such AUVs relying on DL perception for navigation and decision-making, the goal remains to reduce DL model output uncertainties, ensuring the vehicle's and its surroundings' safety.

\section{Conclusion and Future Research Trends}
\label{sec: Conclusion}

The growing interest in underwater exploration, inspection, and monitoring has led to a specific need for underwater data collection and interaction, resulting in underwater robots like AUVs. While AUVs, an acronym for Autonomous Underwater Vehicles, mean autonomy under real-time human operators' supervision, they still seek fully autonomous actions and interactions in the deep sea. In contrast with other autonomous vehicles, such as autonomous cars and  UAVs, AUVs suffer from a lack of communication, visibility, and available data, which results in an uncertain and dangerous environment for vehicles to perform tasks autonomously. Furthermore, autonomous behaviors rely on sensors that capture and understand the vehicle's surroundings to adapt to the vehicle's trajectory in real-time. Because of the often bad quality of underwater camera images, underwater vision mainly relies on sonars (e.g., SSS, FLS, MBES) to map and collect data from the underwater environment. Thus, improving AUVs' autonomy requires understanding the sonar data while surveying, which results in implementing computer vision DL models, such as classification, object detection, segmentation, and SLAM. However, implementing such a model onboard requires a safe DL model without real-time supervision and communication. Thus, the challenge is finding the tradeoff between autonomy and safety to improve the vehicle's autonomy without compromising its safety and the safety of its surroundings. In this paper, we tackled the robustness of the sonar-based DL model with the following question: "How can we rely on a real-time sonar-based deep learning model?" aiming to spotlight the current research topics and method that can be applied to reduce the sonar-based DL models uncertainties. We compared previous surveys on sonar-based DL models, highlighting the need for robustness focus. Indeed, current surveys mostly compare sonar-based DL models and highlight the lack of open-source datasets but without referring to the robustness of the model itself, which is primordial to ensure its good behavior. 
By presenting and critically analyzing 19 sonar datasets and comparing various underwater simulators, we have offered a comprehensive resource for accessing and generating open-source data crucial for advancing DL underwater applications. In addition, to support the need for accessible datasets, we provide a novel GitHub repository that clusters the sonar datasets in a single place, open for community contributions. The discussion on robustness, through the lenses of neural network verification, adversarial attacks, and OOD detection, underscores the need for the resilience of sonar-based DL models. The proposed workflow for enhancing model robustness aims to mitigate uncertainties, resulting in more reliable and safer underwater robotic operations. This increased implementation of onboard DL for underwater missions results in the need for robustness of sonar-based DL models, which will play a pivotal role in ensuring mission success and safety by bridging the gap between theoretical robustness and practical, real-world efficacy. Advancing this frontier will provide new potential in autonomous underwater navigation and data collection.
This paper encourages future research to focus on sonar DL models' uncertainties to improve their robustness. Future work should reduce the gap between sim-to-real for sonar data collection and validation, improving the current SOA datasets by publishing collected data to create a baseline dataset. Finally, the current method for DL robustness, such as neural network verification, adversarial attack, and OOD, in the scope of sonar images require a specific focus to ensure reliable DL predictions and AUV safety.

\vfill

\end{document}